\documentclass[lettersize,journal]{IEEEtran}
\usepackage{amsmath,amsfonts}
\usepackage{algorithmic}
\usepackage{algorithm}
\usepackage{array}
\usepackage[caption=false,font=normalsize,labelfont=sf,textfont=sf]{subfig}
\usepackage{textcomp}
\usepackage{stfloats}
\usepackage{url}
\usepackage{verbatim}
\usepackage{graphicx}
\usepackage{cite}
\usepackage{bbm}
\usepackage{multirow}
\usepackage[normalem]{ulem}
\usepackage[caption=false,font=normalsize,labelfont=sf,textfont=sf]{subfig}
\usepackage{tikz}
\usepackage{amssymb}
\usepackage{pifont}
\newcommand{\cmark}{\ding{51}}%
\newcommand{\xmark}{\ding{55}}%
\usepackage{orcidlink} 
\hypersetup{hidelinks}

\begin{document}

\title{FastSTI: A Fast Conditional Pseudo Numerical Diffusion Model for Spatio-temporal Traffic Data Imputation}

\author{Shaokang Cheng, Nada Osman, Shiru Qu, and Lamberto Ballan \thanks{
 
Shaokang Cheng and Shiru Qu are with the School of Automation, Northwestern Polytechnical University, Xi'an 710072, China. 

Nada Osman and Lamberto Ballan are with the Department of Mathematics ``T. Levi-Civita'', University of Padova, Padua 35131, Italy.}, \IEEEmembership{Senior Member, IEEE}}


\markboth{}
{FastSTI paper - IEEE Transactions template}

\maketitle

\begin{abstract}
High-quality spatiotemporal traffic data is crucial for intelligent transportation systems (ITS) and their data-driven applications. Inevitably, the issue of missing data caused by various disturbances threatens the reliability of data acquisition. Recent studies of diffusion probability models have demonstrated the superiority of deep generative models in imputation tasks by precisely capturing the spatio-temporal correlation of traffic data. One drawback of diffusion models is their slow sampling/denoising process. In this work, we aim to accelerate the imputation process while retaining the performance. We propose a fast conditional diffusion model for spatiotemporal traffic data imputation (FastSTI). To speed up the process yet, obtain better performance, we propose the application of a high-order pseudo-numerical solver. Our method further revs the imputation by introducing a predefined alignment strategy of variance schedule during the sampling process. Evaluating FastSTI on two types of real-world traffic datasets (traffic speed and flow) with different missing data scenarios proves its ability to impute higher-quality samples in only six sampling steps, especially under high missing rates (60\% $\sim$ 90\%). The experimental results illustrate a speed-up of $\textbf{8.3} \times$ faster than the current state-of-the-art model while achieving better performance. 
\end{abstract}


\vspace{1em} 
\noindent This is the accepted version of the paper that will appear in IEEE Transactions on Intelligent Transportation Systems. The final published version may differ from this version. \\
Early Access DOI: \href{https://doi.org/10.1109/TITS.2024.3469240}{10.1109/TITS.2024.3469240}


\section{Introduction}
\IEEEPARstart{S}{patiotemporal} traffic data, acquired by diverse sensing systems, plays a fundamental role in intelligent transportation systems (ITS), since they allow a wide array of applications and decision-making processes~\cite{chen2021low, lei2022bayesian}. 
Nevertheless, the common presence of missing data due to equipment failures and transmission errors negatively affects downstream tasks, such as traffic flow prediction. Missing values can lead city planning authorities to rerun experiments to gather necessary data, resulting in additional budgetary expenses and time delays~\cite{said2021spatiotemporal}. For decades, missing data imputation techniques have been extensively investigated, mostly trying to identify spatial and temporal correlations from observed data and accurately estimate the missing values~\cite{lana2018imputation}.

Focusing on imputing traffic data, early traditional methods, exemplified by statistical \cite{beretta2016nearest, kihoro2013imputation} or classical machine learning (ML) \cite{white2011multiple, shang2018imputation} are trivially simple that they neglect the potentially complex interactions between spatial and temporal correlations of traffic conditions. Conversely, deep learning (DL)-based techniques have proven practicality in approximating complex functions and better mining the spatiotemporal evolution patterns. For example, RNNs and their variants \cite{cao2018brits, cini2021filling} are commonly employed to impute missing values. Still, RNN-based approaches present the limitation of assuming sequential relationships within time-series data \cite{zhang2021missing, Bertugli2021-acvrnn}. Additionally, RNNs cannot exploit parallel processing capabilities and face difficulties directly modeling the interdependence among input data with distinct timestamps.

Among various imputation methods, deep generative models have gained significant popularity. Recently, diffusion models have emerged as the new state-of-the-art method of deep generative models family \cite{ho2020denoising, song2020score, nichol2021improved}, surpassing the long-standing dominance of generative adversarial networks (GANs) in diverse of challenging domains \cite{yang2022diffusion}. Compared with other probabilistic approaches (e.g., VAEs and GANs), diffusion-based imputation offers stable training and models sophisticated data distributions by sufficient denoising steps \cite{lin2023diffusion}. 

Existing works started to apply Denoising Diffusion Probabilistic Model (DDPM) to impute missing traffic data \cite{tashiro2021csdi, alcaraz2022diffusion, liu2023pristi}. However, these approaches involve iterative procedures with several evaluation steps, which can be time-consuming and inefficient in real-time applications. To guarantee the quality of the generated samples and enhance the model's capacity for representation, a number of denoise iterations are necessary for diffusion-based imputation methods, which lead to higher computational processing. For example, PriSTI \cite{liu2023pristi} experiments on the METR-LA dataset show that diffusion models need around 50 denoising steps for satisfactory estimations, taking $\sim 5s$ to impute average five-minute traffic speed data from 207 sensors. This inevitably limits the real-time performance of missing data imputation.

In this work, we design a fast conditional pseudo-numerical diffusion model for spatiotemporal traffic data imputation, where we apply pseudo-numerical methods and a predefined variance schedule to accelerate the inference time while retaining the imputation precision. Our acceleration technology is general and compatible with most diffusion-based models. To the best of our knowledge, FastSTI is the first to allow real-time application of conditional diffusion models in traffic data imputation with minimal computational cost and negligible loss of accuracy. We can brief our contributions as follows:
\begin{enumerate}
    \item We put forward and integrate the high-order pseudo-numerical solvers into a conditional diffusion model to improve traffic data imputation, both accuracy and speed.
   \item To further accelerate the data imputation process, we tune and utilize a variance schedule that derives a short and effective noise schedule, reducing computation time without significant loss of accuracy.
   \item To leverage observational feature knowledge, we use a graph convolution network (GCN) variants to capture the correlation of traffic conditions from both spatial and temporal perspectives. 
   \item Extensive experiments are conducted to evaluate our proposed imputation method, with different data-missing scenarios and high missing rates, confirming the effectiveness of our FastSTI model.
\end{enumerate}

The remainder of this work is organized as follows: Section \ref{sec:2} reviews related work. Section \ref{sec:3} defines the preliminaries. Section \ref{sec:4} introduces our proposed method FastSTI. Section \ref{sec:5} presents experiments and performance discussions, and we conclude in Section \ref{sec:6}.


\section{Related Work} \label{sec:2}
\subsection{Spatiotemporal Traffic Data Imputation}
Spatiotemporal traffic data imputation has emerged as a prominent area of study in urban computing. Early conventional methods mainly relied on statistical or classical machine learning (ML) techniques, overlooking the complex spatiotemporal patterns within the city's road network. These methods include but are not limited to Mean, Linear, KNN, MICE, VAR, and KF \cite{beretta2016nearest, white2011multiple}. Likewise, some studies use matrix and tensor factorization methods, which typically rely on low-order decomposition to estimate or impute missing data \cite{yu2016temporal, chen2019missing}. In recent years, DL-based methods have become widely employed when dealing with traffic data imputation tasks that require complex and nonlinear situations. A common approach in DL is to utilize RNNs and their variants, such as LSTMs and GRUs, for time-series imputation \cite{ming2022multi, cao2018brits, cini2021filling}. Nevertheless, RNN-based imputation models are time-consuming, sensitive to error propagation during imputation, and struggle to capture dynamic changes effectively.

Recent deep generative-based models have made significant advancements in spatiotemporal traffic data imputation, such as variational autoencoders (VAEs) \cite{fortuin2020gp, qin2021network, mattei2019miwae}, generative adversarial networks (GANs) \cite{yoon2018gain, yuan2022stgan}, and normalizing flows \cite{chen2021learning, richardson2020mcflow}. As a state-of-the-art class of deep generative models, diffusion models have emerged as strong contenders challenging the long-standing supremacy of GANs \cite{yang2022diffusion}. For instance, Tashiro et al. \cite{tashiro2021csdi} proposed a probabilistic imputation method to directly learn the conditional distribution with conditional score-based diffusion models. Alcaraz et al. \cite{alcaraz2022diffusion} put forward a combination of state-space models as effective blocks for capturing long-term dependencies in time series with conditional diffusion models. Also, Liu et al. \cite{liu2023pristi} proposed PriSTI, a global context prior diffusion framework for imputing spatiotemporal data. To enhance accuracy, PriSTI incorporates conditional information, spatiotemporal global correlations, and geographic relationships. However, the above methods pose challenges to the practical application of data imputation due to the extensive inference time. In this work, we propose a fast, high-quality traffic data imputation diffusion model.

\subsection{Diffusion Models}
Diffusion models are a family of probabilistic generative models that proved outstanding performance in various domains, including but not limited to computer vision \cite{brempong2022denoising}, spatiotemporal data modeling \cite{tashiro2021csdi}, natural language processing \cite{austin2021structured}, and multi-modal learning\cite{ramesh2022hierarchical}. Ho et al. \cite{ho2020denoising} propose the Denoising Diffusion Probabilistic Model (DDPM), utilizing two Markov chains process of the $T$-steps: 1) a $diffusion \ process$ that perturbs data by adding noise, and 2) a $reverse \ process$ that reconstructs the data from the noise. Given a data distribution $\acute{x} _{0} \sim p(\acute{x} _{0})$, and $\acute{x} _{t}$ is the sampled latent variable sequence, with $t = 1, \cdots , T$ denoting the diffusion steps. The $diffusion \ process$ gradually adds standard Gaussian noise into $\acute{x} _{0}$ until it becomes close to $\acute{x} _{t}$, while the $reverse \ process$ denoises $\acute{x} _{t}$ to recover $\acute{x} _{0}$.

However, the process of generating samples from DDPM requires iterative approaches that involve multiple evaluation steps. Many works have focused on accelerating the sampling process and improving the quality of the resulting samples by considering stochastic differential equations (SDEs) or ordinary differential equations (ODEs). Compared to SDE-slover, ODE-solver brings a more promising approach as the trajectories they solve are deterministic and not influenced by random fluctuations. One example of early work in accelerating diffusion model sampling is Denoising Diffusion Implicit Models (DDIM) \cite{song2020denoising}, considered as a first-order ODE-solver. In extensive experimental investigations, Karras et al. \cite{karras2022elucidating} recently demonstrated that high-order numerical solvers achieve a better trade-off between sample quality and speed. By obtaining the ODE form, many numerical solvers can be readily applied to the sampling process. To a certain extent, high-order solvers reduce discretization errors but also increase computational requirements. Instead, our framework enhances the denoising process of the diffusion model by integrating the variants of high-order numerical solvers and presenting an acceleration method that effectively reduces time consumption.

\subsection{Numerical Methods}
As stated above, various numerical methods can provide high-order approaches for solving ODEs \cite{wanner1996solving}, such as:
\begin{enumerate}
    \item Heun’s Method: also named improved Euler, it modifies Euler’s method from one step into a two-step to improve accuracy.
    \item Runge-Kutta Methods (RK): a class of numerical techniques that enhance accuracy by incorporating information from multiple hidden steps.
    \item Linear Multi-Step Method (LMS): Similar to but different from RK methods, the linear multi-step method utilizes previous steps rather than hidden steps to estimate the next step.
\end{enumerate}

However, \cite{salimans2022progressive, liu2022pseudo} also showed that these classical numerical methods introduce significant noise at a high speedup ratio, causing the solver to sample data that deviates from the main distribution area. To deal with this issue, Liu et al. \cite{liu2022pseudo} designed a pseudo-numerical method for unconditional diffusion models by introducing a nonlinear transfer part as the pseudo-numerical method rather than directly using classical numerical methods. Our approach leverages the pseudo-numerical solvers into a conditional diffusion model to further extract additional prior knowledge, enhancing the imputation accuracy of traffic data.

\renewcommand{\arraystretch}{1.2}
\begin{table}[t]
\caption{Notation\label{tab:Notation}}
\centering
\begin{tabular}{ll}
\hline
Notations                                            & Descriptions                                                 \\ \cline{1-2}
$X$                                                  & Road traffic data                                                 \\
$\Acute{X}$                                          & Manually selected imputation target                          \\
$L$                                                  & The length of time steps                                     \\
$N$                                                  & The number of the observation nodes                          \\
$G$                                                  & Road traffic network                                              \\
${A}$                                                & Adjacency matrix of geographic information                   \\
${K}$                                                & Graph random walk steps                                      \\
${\chi}$                                             & Conditional observations                                     \\
$T$                                                  & The number of diffusion steps                                \\
$\alpha _{t},\beta _{t},\bar{\alpha} _{t},\epsilon$  & Constant hyperparameters of diffusion process                \\
$\epsilon _{\theta} $, $\bar{\varphi}_{t}$, $\bar{\xi}_{t}$    & Constant hyperparameters of imputation process     \\
$T_{acc}$                                            & The number of accelerated denoising steps                    \\ \cline{1-2}
\end{tabular}
\end{table}


\section{Preliminaries} 
\label{sec:3}
Our task is to estimate the missing data or the corresponding distributions in traffic datasets with incomplete observed values. We define the traffic network as a graph $G$, exploit the spatiotemporal correlation within the network, and impute the missing values. Table \ref{tab:Notation} summarizes the notations used throughout our work.

Formally, let $X=\left \{ x_{1},x_{2},\dots ,x_{L}  \right \} $ be a sequence with shape $R ^{L\times N} $, where $L$ represents the length of time steps, and $N$ is the number of observation nodes (e.g., traffic sensors/loop detectors). The observation nodes can be denoted with a directed graph $G= (V, E, A)$, where $V$ and $E$ represent a finite set of nodes and edges, respectively. The adjacency matrix $A \subseteq R^{N \times N}$ is a distance weight matrix, standing for the geographic distance between each pair of observation nodes ($v_{i}$, $v_{j}$). Here, we use threshold Gaussian Kernel \cite{shuman2013emerging} to obtain the adjacency matrix from the geographic distances among nodes.

To account for missing values, we use a binary mask $M_{l} \in \left \{ 0,1 \right \} ^{N}$ that indicates which node of $x_{l}$ are observed in $X$. Specifically, if $m_{l}^{i,j} =0$, the corresponding element $x_{l}^{i,j}$ is missing, while $m_{l}^{i,j} = 1$ denotes that $x_{l}^{i,j}$ is an observed value. 
To handle missing data in practical scenarios where ground truth is unavailable, we manually select imputation targets $\Acute{X}$ from the observed data and then use the binary mask $ \Acute{M_{l} \in \left \{ 0,1 \right \} ^{N}}$ to identify these targets. In this way, we can assess the efficiency of our imputation technique under real-life conditions.


\section{Methodology} 
\label{sec:4}

\begin{figure*}[!ht]
    \centering
    \includegraphics[width=.9\linewidth]{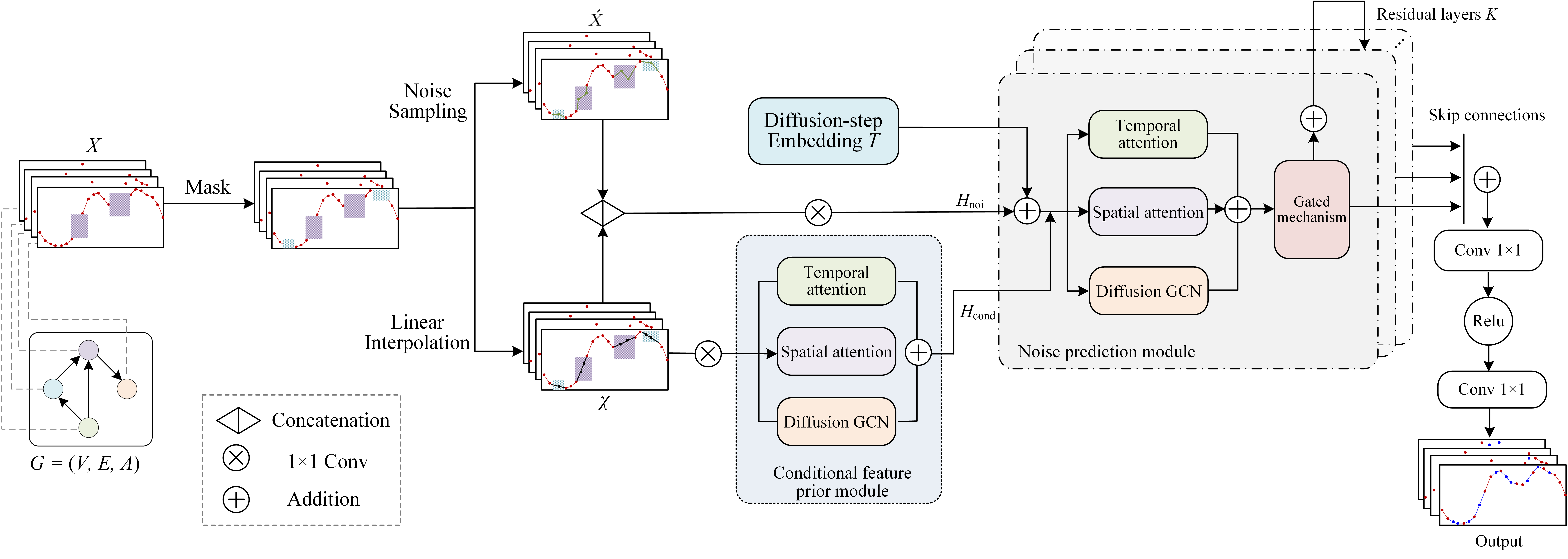}
    \caption{Proposed FastSTI Model Architecture. In FastSTI, we take observed values and geographic location information as input. Our approach uses linear interpolation and leverages the conditional feature prior module to model the prior spatiotemporal context. Afterwards, the prior feature weights are obtained and fed into the noise prediction module to help predict noise.}
    \label{fig:FastSTI pipeline}
\end{figure*}

In this section, we describe the basic ideas of our proposed Fast Conditional Pseudo Numerical Diffusion Model for Spatio-temporal Traffic Data Imputation (FastSTI). Figure \ref{fig:FastSTI pipeline} illustrates the architecture of our proposed method. We build our methodology on top of the state-of-the-art PriSTI model \cite{liu2023pristi}. Different from PriSTI \cite{liu2023pristi}, we propose a high-order conditional pseudo-numerical method to improve the sampling quality in the reverse process of the diffusion model. Meanwhile, we explore an accelerated method by using variance scheduling to address limitations in the inference time of diffusion model-based baselines. In addition, a variant of GCN (Diff-GCN) is adopted to enhance the extraction of local spatial correlations of traffic patterns (i.e., traffic speed and flow), in replacement of the direct message passing approach used in PriSTI.

\subsection{Masking Strategy of Imputation Targets}

Given an observed sequence $X$, we split it into two parts: one represents the imputation target $\Acute{X}$, while the other represents the observed values serving as conditional observations. To simulate different real-life scenarios of missing traffic data, following \cite{cini2021filling}, we consider two masking strategies:
\begin{itemize}
 \item\textbf{Block-missing scenario:} Missing values occur in contiguous blocks over time. We randomly mask $5\%$ of the available data and adopt simulated failures with a probability of $0.15\%$ for each node/sensor. The duration of each failure is sampled uniformly from the interval [$min\_steps, max\_steps$], where $min\_steps$ and $max\_steps$ correspond to the length of time steps.
 
 \item\textbf{Point-missing scenario:} Random occurrence of missing values, where $25\%$ of observations are masked in a random manner.
\end{itemize}

\subsection{Conditional Diffusion Model}

Our conditional diffusion model utilizes the observed values and geographic location data of all sensors as inputs. This input information is processed through a coarse interpolation and conditional feature prior module to model and extract the spatiotemporal correlations of traffic patterns (i.e., traffic speed and flow). Subsequently, the noise prediction module performs reverse denoising using the aforementioned conditional feature information to generate the imputation output.

\textbf{Coarse Interpolation.} To provide rough but efficient interpolated conditional information denoted as $\chi$, Linear interpolation (Lin-TIP) is employed to fill in the missing data at each node. This approach relies on the uniform distribution of missing data to ignore the randomness of time series but retains certain spatiotemporal relations. We can also observe the baseline result of Lin-TIP from Table \ref{tab:Comparing}. Meanwhile, Lin-TIP, a simple architecture, meets the requirements for real-time performance.

\textbf{GCN-based Conditional Feature Prior.} The linear interpolation method assumes uniform and linear changes in traffic states. However, traffic data exhibits dynamic temporal dependencies, and the flow of different regions/interactions affects each other, making linear interpolation inadequate for capturing the nonlinear and random patterns in real-life traffic conditions \cite{yang2021st}. Therefore, it is necessary to utilize a learnable module that can adapt to such non-linearity and randomness in real-life traffic patterns. To address the problem, we adopt a \textit{conditional feature prior module} $ \rho\left ( \cdot \right )$, which takes the interpolated information $\chi$ and the adjacency matrix $\mathcal{A}$ to model the nonlinear conditional information, extracting global spatial and temporal correlations as well as local geographic correlations of road traffic patterns, as illustrated in Eq. \ref{eq:conditonal}, where the input $\mathcal{H}$ denotes a 1$d$ convolution of the interpolated data ($\mathcal{H}=Conv(\chi)$). Similar to \cite{liu2023pristi}, the global temporal correlation is captured using a transformer-based self-attention module $\phi _{Tem}\left ( \mathcal{H} \right )$ (Eq. \ref{eq:Tem}), while another global self-attention module $\phi_{Spa}\left ( \mathcal{H} \right )$ extracts the global spatial correlations, as in Eq. \ref{eq:Spa}. To better model the local geographic correlations, our conditional prior module replaces the message-passing network in \cite{liu2023pristi} with a graph convolution module $\phi _{\mathrm{DGCN}}\left ( \mathcal{H, A} \right ))$, as in Eq. \ref{eq:DGCN}.

\begin{equation}
\label{eq:conditonal}
\rho\left ( \mathcal{H, A} \right ) =  \mathrm{MLP}( \phi _{Tem}\left ( \mathcal{H} \right ) + \phi_{Spa}\left ( \mathcal{H} \right )+\phi _{\mathrm{DGCN}}\left ( \mathcal{H, A} \right ))
\end{equation}

\begin{equation}
\phi _{Tem}\left ( \mathcal{H} \right ) = \mathrm{Norm}\left ( Attn_{Tem}(\mathcal{H}) + \mathcal{H}\right ) 
\label{eq:Tem}
\end{equation}
\begin{equation}
\phi _{Spa}\left ( \mathcal{H} \right ) = \mathrm{Norm}\left ( Attn_{Spa}(\mathcal{H}) + \mathcal{H}\right ) 
\label{eq:Spa}
\end{equation}
\begin{equation}
\phi _\mathrm{{DGCN}}\left ( \mathcal{H, A} \right ) =\mathrm{Norm}\left ( \mathrm{DiffGCN}(\mathcal{H, A}) + \mathcal{H}\right ) 
\label{eq:DGCN}
\end{equation}

In contrast to the message-passing network (MP) employed in PriSTI\cite{liu2023pristi}, our framework captures the local geographic correlations with an improved graph convolution network module named Diffusion-GCN (Diff-GCN). The local geospatial dependency in traffic flow is modeled by correlating it with a diffusion flow process, described by \cite{li2017diffusion}, which effectively extracts the random nature of traffic dynamics, in addition to featuring a more lightweight structure during the diffusion process, meeting real-time requirements. Additionally, Diff-GCN benefits from a bi-directional random walk $K$ strategy, providing enhanced flexibility in capturing influences from upstream and downstream traffic conditions (e.g., speed or flow).

Specifically, the key operation of diffusion graph convolution over the coarse interpolated data $\chi$ is defined in Eq. \ref{eq:DiffGCN}, where $A$ denotes an adjacency matrix (road distance-based node matrix), $K$ is the step of graph random walk, $\varrho \in [0,1]$ is the graph coefficient, $\theta _{k} $ is the parameter of the convolutional filter, and $ D_{g}$ and $D_{c}$ are the transition matrices of the graph diffusion process and the converse one. Here, $D=diag(A*1)$, matrix $1 \in R^{N}$  represents all element is one.

\begin{equation}
\mathrm{DiffGCN}(\mathcal{H, A})= \sum_{k=0}^{K} \left ( \varrho _{k\ge 1}\left ( \theta _{k}^{1}   \left ( D_{g} A \right )^{k} + \theta _{k}^{2}  \left ( D_{r} A \right )^{k}\right )  \right ) \chi  
\label{eq:DiffGCN}
\end{equation}

Simply put, Diff-GCN generates a group of node features as output after aggregating graph information from the self-node and its $K$-hop neighbors. Notably, the number of neighbors in the graph random walk $K$ is a hyperparameter to be tuned, where a large $K$ value can enable graph convolution to capture more geospatial information. The learning process also becomes more complex and the computation time increases when larger convolution filters are chosen.

Once the conditional feature prior module $H_{\mathrm{cond}}$ is established, these coarse yet useful conditional features are fed into the noise prediction module to facilitate the learning of spatiotemporal correlations of road traffic patterns.

\textbf{Noise Prediction Module.} The noise prediction module is designed to utilize conditional information to predict the missing values, as shown in Fig. \ref{fig:FastSTI pipeline}. The module takes two inputs: 1) The conditional prior $H_{cond}$; 2) Noise information $H_{noi} = Conv(\chi || \Acute{X})$, where $Conv(.)$ is the 1$d$ convolution, $(||)$ represents concatenation, and $\Acute{X}$ is the data sequence with the missing data sampled from a standard Gaussian noise. The temporal, spatial, and geographic modules of the noise prediction are the same modules of the conditional feature prior. However, the noise information $H_{noi}$ would not provide an accurate representation of real-life traffic data; therefore, the conditional information $H_{cond}$ is used for both the query and value of the temporal attention $\phi _{Tem}$ and spatial attention $\phi_{Spa}$, while  $H_{noi}$ is used for the key.

The output of each layer in the noise prediction module is split into a residual connection and skip connections. The residual connection is the input of the next layer, and the skip connections of each layer are added and fed into a two-layer 1$d$ convolution to obtain the output of the noise prediction module, where the output contains only the value of the imputation target.

\subsection{Training Procedure}
Given an observed sequence $X$ and split into two parts: one part represents the imputation target $\Acute{X}$ which is generated by the masking strategies (i.e., block-missing and point-missing strategy), while the remaining observed values serve as conditional observations ${\chi}$.

Next, given the imputation target $\Acute{X}$ and the interpolated conditional observations ${\chi}$, we sample the imputation target $\Acute{X}$ and train the noise prediction model $\epsilon_{\theta}$ by minimizing the loss function. Where the noise predictor minimizes the loss function $\mathcal L_{t}$, defined in Eq. \ref{eq:loss}, where the imputation target $\Acute{X}_{t} =\sqrt{\bar{\alpha}_{t}}x_{0}+\sqrt{1-\bar{\alpha}_{t}}\epsilon$; $\bar{\alpha}_{t}= \prod_{i=1}^{t} \alpha_{i}$, $\alpha_{t}=1-\beta_{t}$. The training procedure is summarized in Algorithm \ref{alg:alg1}.

\begin{equation}
     \mathrm{min}{{\mathcal L_{t}= \mathrm{min} \mathbb{E}_{\Acute{X}\sim q(\Acute{X}_{0}),\epsilon \sim N(0,I)}   \left |  \right | \boldsymbol \epsilon -\boldsymbol \epsilon _{\theta } } (\Acute{X}_{t}  , {\chi} ,A,t)\left |  \right | ^{2}}
    \label{eq:loss}
\end{equation}

\begin{algorithm}[!t]
\caption{Training procedure}\label{alg:alg1}
\begin{algorithmic}
\STATE 
\STATE  $\textbf{Require:} \ \mathrm{incomplete \ observed \ data} \ X, \mathrm{the \ noise \ levels} \ \bar{\alpha} _{t},$\STATE $\mathrm{the \ adjacency \ matrix} \ A, \mathrm{the \ number \ of \ iteration} \ N_{i}^{t},$
\STATE  $\mathrm{the \ diffusion \ steps } \ T. $
\STATE  $\textbf{for } {i=1} {\ \rm to} {\ N_{i}^{t}}  \textbf{ do }$
\STATE \hspace{0.5cm}$ \rm{Sample} \ {x_{T} \sim q_{data}}, \boldsymbol \epsilon \sim \left(0, \boldsymbol{I} \right)$,
\STATE \hspace{0.5cm}$ {t \sim {\rm Uniform} \left ( \left \{ 0, \dots, {T-1} \right \}  \right )} $
\STATE \hspace{0.5cm}$ \rm{Compute} \ \Acute{X}_{t} =\sqrt{\bar{\alpha}_{t}}x_{0}+\sqrt{1-\bar{\alpha}_{t}}\epsilon $
\STATE \hspace{0.5cm}$ {{\mathcal L_{t}=\left |  \right | \boldsymbol \epsilon -\boldsymbol \epsilon _{\theta } } (\Acute{X}_{t}  , {\chi} ,A,t)\left |  \right | ^{2}} $
\STATE \hspace{0.5cm}$ {\rm Take\ \rm gradient\ \rm descent\ \rm step\ \rm on \ \rm \nabla_{\theta }  \mathcal L_{t}} $
\STATE  $\textbf{end for}$
\end{algorithmic}
\label{alg1}
\end{algorithm}

\subsection{Conditional Pseudo-Numerical Methods}
As described earlier, we aim to speed up the imputation process and improve its quality. Inspired by \cite{liu2022pseudo}, we introduce the pseudo-numerical solvers to our conditional diffusion model during the reverse/sampling process. Different from the baseline models PriSTI \cite{liu2023pristi} and CSDI \cite{tashiro2021csdi} that directly use reverse sampling calculation of DDPM, our approach leverages the higher-order numerical solver to sample the distribution of data and avoid the influence of random noise caused by DDPM, thereby enhancing the quality of data generation.

In this work, we apply two kinds of pseudo-numerical methods for the conditional diffusion model in our imputation task: \textbf{FastSTI-2} ($2^{nd}$-order) and \textbf{FastSTI-4} ($4^{th}$-order). As DDPM can be considered a special case of DDIM \cite{song2020denoising}, we provide the following steps to develop the high-order pseudo-numerical sampling method in the conditional diffusion model. First, the reverse process of DDIM is transformed into an ordinary differential equation (ODE) form. Then, the nonlinear transfer part is combined with three classical numerical techniques, namely Heun's methods, Runge-Kutta methods (RK), and Linear Multi-Step methods (LMS). Meanwhile, the conditional information is built into the pseudo-numerical diffusion model for our task of traffic data imputation.

Mathematically, the reverse process of DDIM is defined in Eq. (\ref{eq:t_{DDIM}}). To construct the ODE form of the diffusion model, the equation is reformulated by subtracting $x_{t}$ from both sides. This reformulation makes it equivalent to a numerical step in solving the ODE, as demonstrated in Eq. (\ref{eq:ODE_{DDIM}}).

Then, the discrete-time step $t-1$ in Eq. (\ref{eq:t_{DDIM}}) is replaced with a continuous version represented by $t-\Delta t$, and by letting $\Delta t$ tends to $0$, the ODE becomes:

\begin{equation}
    \frac{d x}{d t}=-\bar{\alpha}^{\prime}(t)\left(\frac{x(t)}{2 \bar{\alpha}(t)}-\frac{\epsilon_{\theta}(x(t), t)}{2 \bar{\alpha}(t) \sqrt{1-\bar{\alpha}(t)}}\right) 
    \label{eq:ode_derivative}
\end{equation}

After obtaining the ODE, the pseudo-numerical method categorizes the classical numerical methods (Heun’s, RK, and LMS) into two components:

\begin{enumerate}
    \item \textit{The gradient part}, responsible for determining the gradient at each step, e.g., $4^{th}$-order linear multi-steps gradient part $ {f}' = \frac{\Delta t{}}{24}(55f_{t}-59f_{t-\Delta t}+37f_{t-2\Delta t}-9f_{t-3\Delta t})$.
    \item \textit{The transfer part}, as shown in Eq. (\ref{eq:transfer}), which generates the result for the next step, i.e., $x_{t-\Delta t} = x_{t}+ \Delta t {f}'$.
\end{enumerate}

All numerical methods share the same transfer part, while their gradient parts differ. The transfer part $x_{t-\Delta t}$ is obtained by $\nu$, rewriting Eq. (\ref{eq:ODE_{DDIM}}) into Eq. (\ref{eq:transfer}), where ${\chi}$ represents the conditional observations. While the gradient part is defined for each numerical method in Table \ref{tab:grad_eqs}.

\begin{equation}
\begin{split}
 & \nu \left(x_{t}, \epsilon_{t}, \chi ,A,t,t-\Delta t \right)=  \\
 & \frac{\sqrt{\bar{\alpha}_{t-\Delta t}}}{\sqrt{\bar{\alpha}_{t}}} x_{t}- \frac{\left(\bar{\alpha}_{t-\Delta t}-\bar{\alpha}_{t}\right)}{\sqrt{\bar{\alpha}_{t}}\left(\sqrt{\left(1-\bar{\alpha}_{t-\Delta t}\right) \bar{\alpha}_{t}}+\sqrt{\left(1-\bar{\alpha}_{t}\right) \bar{\alpha}_{t-\Delta t}}\right)} \epsilon_{t}
\end{split}
\label{eq:transfer}
\end{equation}

\begin{algorithm}[!t]
\caption{Imputation (sampling) process of FastSTI-4}\label{alg:alg2}
\begin{algorithmic}
\STATE 
\STATE  $\textbf{Require:} \ \mathrm{observed \ data} \ X, \mathrm{adjacency \ matrix} \ A,$
\STATE $ \mathrm{\# iterations} \ N_{i}^{t}, \mathrm{noise \ predictor} \ \epsilon _{\theta }, \mathrm{diffusion \ steps } \ T. $
\STATE  $x_{T} \sim  \mathcal N\left(0, \boldsymbol{I} \right)$
\STATE  $\textbf{for } {t=T-1, T-2, T-3} \textbf{ do }$
\STATE \hspace{0.5cm}$ {x_{t}, e_{t} = \mathrm{PRK4} (x_{t+1}, \chi, A, t+1,t)} $
\STATE  $\textbf{end for}$
\STATE  $\textbf{for } {t=T-4, \dots, 0} \textbf{ do }$
\STATE \hspace{0.5cm}$ {x_{t}, e_{t} = \mathrm{PLMS4} (x_{t+1}, {\left \{ e_{p} \right \} } _{p > t},\chi, A, t+1, t)} $
\STATE  $\textbf{end for}$
\STATE  $\textbf{return } {x_{0}} $
\end{algorithmic}
\label{alg2}
\end{algorithm}

Algorithm \ref{alg:alg2} shows an example of our FastSTI-4, where we initially adopt the 4th-order pseudo Runge-Kutta (PRK4) method to obtain the results of the first three steps, followed by the utilization of the $4^{th}$-order pseudo-linear multi-step method (PLMS4) to compute the remaining. Similarly, FastSTI-2 employs the 2nd-order pseudo Heun’s (PH2) to obtain the results of the first two steps, followed by utilizing the $2^{th}$-order pseudo-linear multi-step method (PLMS2) to calculate the remaining. Our conditional pseudo-numerical methods are compatible with many diffusion-based models (e.g., PriSTI \cite{liu2023pristi}). We leverage one of the sampling methods to the baseline model in the experiments (see Sec. \ref{sec:8}).

\begin{equation}
    x_{t-1}=\sqrt{\frac{\bar{\alpha}_{t-1}}{\bar{\alpha}_{t}}}\left (  x-\sqrt{1-\bar{\alpha}_{t}}\epsilon _{\theta}(x_{t},t )  \right )+ \sqrt{1-\bar{\alpha}_{t-1}} \epsilon _{\theta}(x_{t},t ) 
    \label{eq:t_{DDIM}}
\end{equation}
\begin{equation}
\begin{split}
    & x_{t-\Delta t}-x_{t} = \left(\bar{\alpha}_{t-\Delta t}-\bar{\alpha}_{t}\right)\left(\frac{x_{t}}{\sqrt{\bar{\alpha}_{t}}\left(\sqrt{\bar{\alpha}_{t-\Delta t}}+\sqrt{\bar{\alpha}_{t}}\right)} \right. \\
    & \quad \left. -\frac{\epsilon_{\theta}\left(x_{t}, t\right)}{\sqrt{\bar{\alpha}_{t}}\left(\sqrt{\left(1-\bar{\alpha}_{t-\Delta t}\right) \bar{\alpha}_{t}}+\sqrt{\left(1-\bar{\alpha}_{t}\right) \bar{\alpha}_{t-\Delta t}}\right)}\right)
\end{split}
\label{eq:ODE_{DDIM}}
\end{equation}

\begin{table}[]
\centering
\caption{Gradient equations of the different pseudo numerical methods}
\label{tab:grad_eqs}
\begin{tabular}{l}
\hline
$2^{nd}$-order pseudo linear multi-step (PLMS2)\\$\begin{aligned}[t]
     \setlength{\jot}{1pt}
     \left\{\begin{aligned}
  & \  e_{t}  = \epsilon _{\theta } \left( \acute{x}_{t},\chi ,A,t  \right ) , \ \ \ \ \ \ \ \ \ \ \ \ \ \ \  \ \   \\ 
  & \  {e}' _{t} = \frac{{1}}{2}(3e_{t}-e_{t-\Delta t}), \ \ \\ 
  & \  x_{t-\Delta t} = \nu (\acute{x}_{t},\chi ,e_{t}^{'},A,t,t-\Delta t ).
\end{aligned}\right. 
    \label{eq:PLMS2}
\end{aligned}$\\
\hline
$2^{nd}$-order pseudo Heun's (PH2)\\$\begin{aligned}[t]
    \setlength{\jot}{1pt}
    \left\{\begin{aligned}
    & \ e_{t}^{1}   = \epsilon _{\theta } \left( \acute{x}_{t},\chi ,A,t  \right ),\\
    & \ x_{t}^{1} = \nu (\acute{x}_{t},\chi ,e_{t}^{1},A,t,t-\Delta t ),\\
    & \ e_{t}^{2}   = \epsilon _{\theta } \left( \acute{x}_{t}^{1},\chi ,A,t-\Delta t \right ),\\
    & \ {e}' _{t}  = \frac{1}{2} (e_{t}^{1}+e_{t}^{2}),\\
    & \ x_{t-\Delta t} = \nu (\acute{x}_{t},\chi ,e_{t}^{'},A,t,t-\Delta t ).
    \end{aligned}\right. 
    \label{eq:HS}
\end{aligned}$\\
\hline
$4^{th}$-order pseudo linear multi-step (PLMS4)\\$\begin{aligned}[t]
     \setlength{\jot}{1pt}
     \left\{\begin{aligned}
  & \  e_{t}  = \epsilon _{\theta } \left( \acute{x}_{t},\chi ,A,t  \right ) , \ \ \ \ \ \ \ \ \ \ \ \ \ \ \  \ \   \\ 
  & \  {e}' _{t} = \frac{{1}}{24}(55e_{t}-59e_{t-\Delta t}+37e_{t-2\Delta t}-9e_{t-3\Delta t}), \ \ \\ 
  & \  x_{t-\Delta t} = \nu (\acute{x}_{t},\chi ,e_{t}^{'},A,t,t-\Delta t ).
\end{aligned}\right. 
    \label{eq:PLMS4}
\end{aligned}$\\
\hline
$4^{th}$-order pseudo Runge-Kutta (PRK4)\\$\begin{aligned}[t]
      \setlength{\jot}{1pt}
      \left\{\begin{aligned}
     & \   e_{t}^{1}   = \epsilon _{\theta } \left( \acute{x}_{t},\chi ,A,t  \right ), \\
     & \   x_{t}^{1} = \nu (\acute{x}_{t},\chi ,e_{t}^{1},A,t,t-\frac{\Delta t }{2} ),\\
     & \   e_{t}^{2}   = \epsilon _{\theta } ( \acute{x}_{t}^{1},\chi ,A,t-\frac{\Delta t }{2} ),\\
     & \   x_{t}^{2} = \nu (\acute{x}_{t},\chi ,e_{t}^{2},A,t,t-\frac{\Delta t }{2} ),\\
     & \   e_{t}^{3}   = \epsilon _{\theta } ( \acute{x}_{t}^{2},\chi ,A,t-\frac{\Delta t }{2} ),\\
     & \   x_{t}^{3} = \nu (\acute{x}_{t},\chi ,e_{t}^{3},A,t,t-\Delta t ),\\
     & \   e_{t}^{4}   = \epsilon _{\theta } \left( \acute{x}_{t}^{3},\chi , A, t-\Delta t   \right ),\\
     & \   {e}' _{t}  = \frac{1}{6} (e_{t}^{1}+2e_{t}^{2}+2e_{t}^{3}+e_{t}^{4}),\\
     & \   x_{t-\Delta t} = \nu (\acute{x}_{t},\chi ,e_{t}^{'},A,t,t-\Delta t ).
        \end{aligned}\right. 
    \label{eq:PRK}
\end{aligned}$\\
\hline
\end{tabular}
\end{table}

\subsection{Accelerated Imputation.} \label{sec:10}
\begin{figure}[t]
    \centering    \includegraphics[width=1\linewidth]{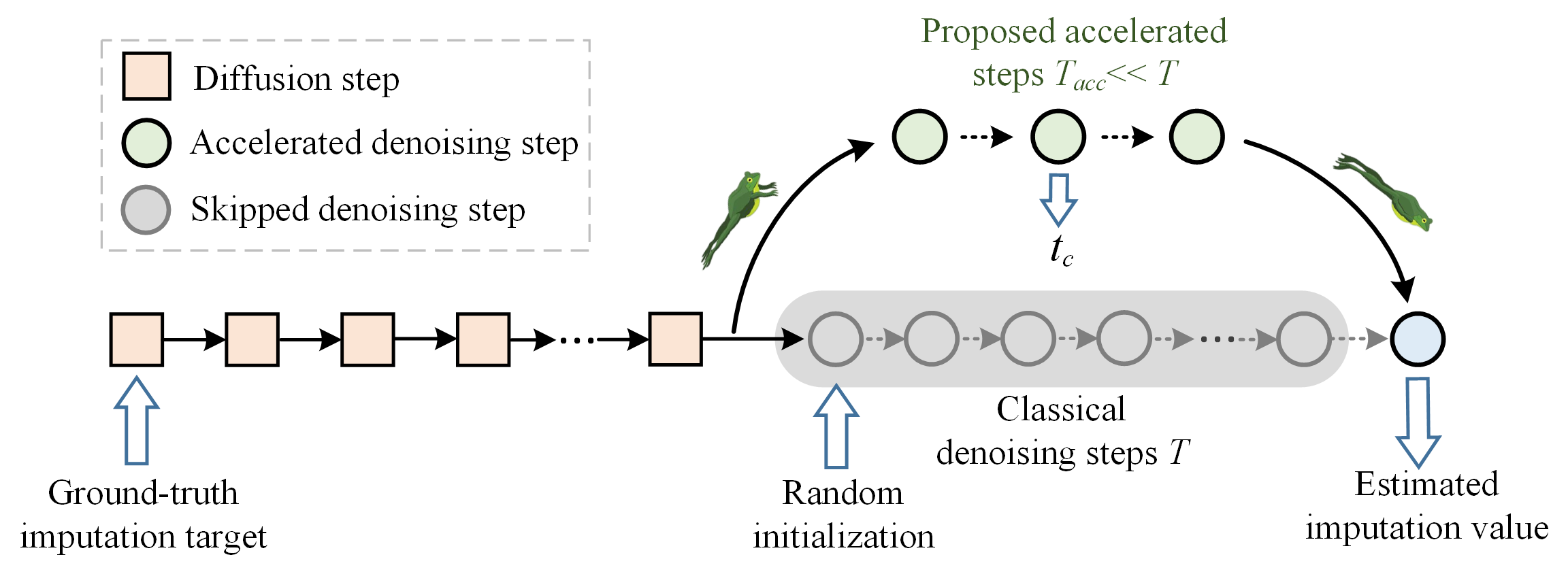}
    \caption{Accelerated Imputation. FastSTI utilizes "schedule alignment" to estimate the denoised distribution, replacing multiple classical denoising steps, thereby accelerating inference without significant loss of accuracy.}
    \label{fig: Accelerated imputation of FastSTI.}
\end{figure}

The application of the reverse process of DDPM for denoising suffers from the drawback of being time-consuming due to the necessity for numerous denoising steps to ensure the quality of the generated samples, see the classical denoising steps (gray part) of Fig. \ref{fig: Accelerated imputation of FastSTI.}. Some typical diffusion-based baselines, such as PriSTI \cite{liu2023pristi} and CSDI \cite{tashiro2021csdi}, with 50 reverse steps, and SSSD \cite{alcaraz2022diffusion}, with 200 steps, use DDPM's sampling way. These baseline models cannot meet real-time requirements in real-world applications.

To accelerate the imputation (sampling) process, inspired by \cite{kong2020diffwave}, we introduce a "schedule alignment" approach that utilizes a predefined number of $T_{acc}$-steps to minimize the imputation time without significant loss of quality. As shown in Fig. \ref{fig: Accelerated imputation of FastSTI.}, the key concept is to align the original $T$-steps reverse process into a condensed $T_{acc}$-steps process using a predefined variance schedule. In this way, our method is able to skip several reverse steps to reduce denoising iterations. Algorithm \ref{alg:alg3} summarizes the proposed accelerated sampling (imputation) procedure.

Formally, given the steps of $T_{acc} \ll T$  in the imputation process and a predefined variance schedule $\left \{ \xi _{t}  \right \} _{t=1}^{T_{acc}}$. Different from the training variance schedule $\left \{ \beta _{t}  \right \} _{t=1}^{T}$, the $\left \{ \xi _{t}  \right \} _{t=1}^{T_{acc}}$, with a manual setting of the values, we can calculate the corresponding constants as follows:
 \begin{equation}
    {\varphi}_{t}=1-\xi_{t}, \
    \bar{{\varphi}}_{t}=\prod_{s=1}^{t} \varphi_{c}, \
    \widetilde{\xi}_{t} =\frac{1-\bar{\varphi}_{t-1}}{1-\bar{\varphi}_{t}} \xi_{t}
    \label{eq:bar_xi}
\end{equation}

Our objective is to determine and interpolate the value of $\sqrt{\bar{{\varphi}}_{c}}$ between the training noise levels: $\sqrt{\bar{\alpha}_{t}}$ and $\sqrt{\bar{\alpha}_{t+1}}$, such that $\sqrt{\bar{{\varphi}}_{c}}$ closely approximates $\sqrt{\bar{\alpha}_{t}}$. Afterward, we obtain the aligned diffusion step $t$ by calculating the floating-point $t_{c}$, as in Eq. (\ref{eq:t_{c}}).

\begin{equation}
    t_{c} = t + \frac{\sqrt{\bar\alpha_{t}}-\sqrt{\bar{{\varphi}}_{c}}}{\sqrt{\bar\alpha_{t}}-\sqrt{\bar\alpha_{t+1}}}
    \label{eq:t_{c}}
\end{equation}

\begin{algorithm}[t]
\caption{Accelerated imputation of FastSTI-4}\label{alg:alg3}
\begin{algorithmic}
\STATE 
\STATE  $ \mathrm{Sample} \ x_{T_{acc} } \sim \mathcal N\left(0, \boldsymbol{I} \right)$
\STATE  $\textbf{for } {c=T_{acc}-1, T_{acc}-2, T_{acc}-3} \textbf{ do }$
\STATE \hspace{0.5cm}$ {x_{c}, e_{c} = \mathrm{PRK4} (x_{c+1}, \chi, A, c+1,c)} $
\STATE  $\textbf{end for}$
\STATE  $\textbf{for } {c=T_{acc}-4, \dots, 0} \textbf{ do }$
\STATE \hspace{0.5cm}$ {x_{c}, e_{c} = \mathrm{PLMS4} (x_{c+1}, {\left \{ e_{p} \right \} } _{p > c},\chi, A, c+1, c)} $
\STATE  $\textbf{end for}$
\STATE  $\textbf{return } {x_{0}} $
\end{algorithmic}
\label{alg3}
\end{algorithm}

It is worth highlighting that our acceleration approach is general and compatible with most diffusion-based models, including PriSTI \cite{liu2023pristi}. To prove this point, we also apply the proposed acceleration technique to the baseline model in the experiments (see Section \ref{sec:9}).


\section{Experiments} 
\label{sec:5}
\subsection{Datasets}
We conduct our experiments on two types of traffic imputation tasks: traffic speed and traffic flow imputation. Table \ref{tab:datasets} describes the statistics of these publicly available real-life datasets. Further details about datasets are listed as follows:

1) \textit{Traffic Speed Data}: METR-LA and PEMS-BAY \cite{li2017diffusion}. The traffic speed data is aggregated and reported by the PeMS system at every 5-minute time intervals. METR-LA comprises four months of traffic speed statistics collected from 207 loop detectors installed in the highway of Los Angeles County. Similarly, PEMS-BAY holds six months of traffic speed data from 325 sensors in the Bay Area. 

2) \textit{Traffic Flow Data}: PEMS04 and PEMS08 \cite{guo2019attention}. PEMS04 includes two months of traffic flow data collected from 307 detectors installed in the San Francisco Bay Area. PEMS08 contains two months of traffic flow data from 170 sensors in San Bernardino. Both datasets are aggregated every 5 minutes.

\renewcommand{\arraystretch}{1.2}
\begin{table}[t]
\setlength{\tabcolsep}{2.5pt}
\centering
     \caption{Statistics of datasets} \label{tab:datasets}
\begin{tabular}{ccccc}
\hline
\multicolumn{1}{l}{} & \multicolumn{2}{c}{Traffic Speed}                                                                                                       & \multicolumn{2}{c}{Traffic Flow}                                                                                                        \\ \hline
Datasets                    & METR-LA                      & PEMS-BAY                           & PeMS04                                     & PeMS08                \\ \hline
\textit{Nodes (Sensors)}      & 207                           & 325                             & 307                                        & 170                    \\
\textit{Edges}                & 1515                           & 2694                           & 340                                        & 275                    \\
\textit{Total Timestamps}     & 34272                   & 52116                      & 16992                          & 17856                                         \\
\textit{Time Span}            & \begin{tabular}[c]{@{}c@{}}01/03/2012 - \\ 27/06/2012\end{tabular} & \begin{tabular}[c]{@{}c@{}}01/01/2017 - \\ 30/06/2017\end{tabular} & \begin{tabular}[c]{@{}c@{}}01/01/2018 - \\ 28/02/2018\end{tabular} & \begin{tabular}[c]{@{}c@{}}01/07/2016 - \\ 31/08/2016\end{tabular} \\
\textit{Time Interval}        & \multicolumn{4}{c}{$5$ minutes}                                                                                                                                                                                                                                                            \\
\textit{Daily Range}          & \multicolumn{4}{c}{00:00 - 24:00}                                                \\ \hline
\end{tabular}
\label{tab:datasets}
\end{table}

\subsection{Evaluation Metrics}

The data sets are split into $70\%$ for training, $10\%$ for validation, and $20\%$ for testing. The evaluation metrics adopted are Mean Absolute Error (MAE), as in Eq. (\ref{eq:mae}); Mean Square Error (MSE)/Root Mean Square Error(RMSE), described in Eq. (\ref{eq:mse}) and Eq. (\ref{eq:rmse}), respectively; and Continuous Ranked Probability Score (CRPS) \cite{matheson1976scoring}. MAE and MSE/RMSE assess the errors between targets and imputed values, with conditional masking $(m_{eval})$ applied to $x_i$ and $\Acute{x}_i$.
 
\begin{equation}
    \mathrm{MAE} = \frac{1}{n} \sum_{i=1}^n | (x_i - \Acute{x}_i)\odot m_{eval}|
    \label{eq:mae}
\end{equation}
\begin{equation}
    \mathrm{MSE} = \frac{1}{n} \sum_{i=1}^n ((x_i - \Acute{x}_i) \odot m_{eval})^2
    \label{eq:mse}
\end{equation}
\begin{equation}
    \mathrm{RMSE} = \sqrt{\frac{1}{n} \sum_{i=1}^n ((x_i - \Acute{x}_i) \odot m_{eval})^2}
    \label{eq:rmse}
\end{equation}

CRPS reflects the compatibility of an estimated probability distribution $P$ with the observed values $x$. CRPS Eq. (\ref{eq:cprs}) is defined as the integral of the quantile loss $\Lambda _{\omega} (P^{-1}(\omega) , x)=(\omega-\mathbbm{1}_{x<P^{-1}(\omega)})(x-P^{-1}(\omega))$, where $ P^{-1}(\omega) $ is the $\omega-$quantile of distribution $P$, $\omega \in [0,1]$ represents the quantile levels, and $\mathbbm{1}$ is the indicator function. Following the same setting in \cite{tashiro2021csdi}, 100 samples are generated to approximate the distribution of missing values. We compute quantile losses for discretized quantile levels with 0.05 ticks, defined in Eq. (\ref{eq:cprs_100}), defining $\mathrm{CRPS}(P, \widetilde{X} )$ as the average CRPS at each imputed point as in Eq. (\ref{eq:cprs_average}).

\begin{equation}
    \mathrm{CRPS}(P^{-1}, x)=\int_{0}^{1}2 \Lambda _{\omega} (P^{-1}(\omega),x)d{\omega}
    \label{eq:cprs}
\end{equation}
\begin{equation}
    \mathrm{CRPS}(P^{-1}, x)\simeq \sum_{i=1}^{19} 2 \Lambda _{i*0.05} (P^{-1}(i*0.05),x)/19
    \label{eq:cprs_100}
\end{equation}
\begin{equation}
    \mathrm{CRPS}(P, \widetilde{X} ) = \frac{ {\textstyle \sum_{\widetilde{x}\in \widetilde{X}}\mathrm{CRPS}(P^{-1} , \widetilde{x})} }{\left | \widetilde{X} \right | } 
    \label{eq:cprs_average}
\end{equation}  

\subsection{Implementation Details}

As explained in our methodology, we apply two missing data scenarios: 1) Block-missing, masking $5\%$ of data in the range of  $[12, 48]$ time steps, with a node-failure probability of $15\%$; 2) Point-missing, masking randomly $25\%$ of the points. Noting that all datasets include initially missing values (i.e., 8.10\% in METR-LA, 0.02\% in PEMS-BAY, 1.59\% in PEMS04, and 0.35\% in PEMS08), except for with manually simulated anomalies. Performance evaluation is conducted primarily on the manually masked test set.

Table \ref{tab:FastSTI hyperparameters} summarizes our hyperparameters, where the model is trained for $200$ epochs with a batch size of $16$, learning rate of $10^{-3}$, and Adam optimizer. The diffusion model noise schedule is Quadratic, and we utilize user-defined variance schedules $\left \{ 0.0001, 0.001, 0.2, 0.3, 0.5, 0.9 \right \}$.

\renewcommand{\arraystretch}{1.2}
\begin{table}[!t]
\caption{FastSTI hyperparameters\label{tab:FastSTI hyperparameters}}
\centering
\begin{tabular}{lc}
\hline
Hyperparameter & Value\\
\hline
Epochs & 200\\
Batch size & 16\\
Sequence length $L$ & 24\\
Learning rate & $1\times 10^{-3}$\\
Weight decay & $1\times 10^{-6}$\\
Residual layers & 4\\
Residual channels $r$ & 64\\
Self-attention heads $h$ & 8\\
Temporal embedding dim $m$ & 128\\
Graph random walk step $K$ & 2\\
Graph coefficient $\varrho$ & 0.1\\
\hline
Diffusion Schedule & Quadratic\\
The minimum noise level $\beta _{1} $ & 0.0001\\
The maximum noise level $\beta _{T} $ & 0.2\\
\hline
Diffusion steps $T$ & 50\\
Accelerated denoising steps $T_{acc}$ & 6 \\
Variance Schedule & $\left \{ 0.0001, 0.001, 0.2, 0.3, 0.5, 0.9 \right \}$ \\
\hline
\end{tabular}
\end{table}

All the experiments are conducted on Intel(R) Xeon(R) W-2133 CPU @3.60GHz and NVIDIA GeForce 3080Ti GPU 12GB. We implement our FastSTI model in Python 3.7 using Pytorch 1.13.0.

\subsection{Baselines}
We compare our FastSTI model with seventeen baseline models, including statistical (Mean, KNN, Linear InTerPolation (Lin-LTP)), classical ML (MICE, Vector AutoRegression (VAR), Kalman Filter (KF)), low-matrix factorization (TRMF, BATF), deep autoregressive (BRITS, GRIN), and deep generative models (V-RIN, GP-VAE, rGAIN, CSDI, SSSD, PriSTI) in the missing data imputation domain. We provide a concise overview of the baseline models as follows: 
 
(1) Mean: using node-level average to impute missing values. (2) KNN \cite{beretta2016nearest}: estimating missing values with observation values from the nearest neighbors based on a similarity measure. (3) Lin-LTP: performing linear interpolation on the time-series data for each node. (4) KF: using Kalman filtering for imputing missing values in temporal observations. (5) MICE \cite{white2011multiple}: a multiple imputation model with chained equations. (6) VAR: a vector autoregressive single-step-ahead predictor. (7) TRMF \cite{yu2016temporal}: a temporal regularized matrix factorization framework for imputation. (8) BATF \cite{chen2019missing}: a Bayesian augmented tensor factorization model for imputing missing traffic data. (9) BRITS \cite{cao2018brits}: a method that uses a bidirectional RNN to impute missing values. (10) GRIN \cite{cini2021filling}: a framework for multivariate time series imputation by bidirectional graph recurrent neural network. (11) V-RIN \cite{mulyadi2021uncertainty}: a variational autoencoder framework with a recurrent neural network and a modified loss function to impute the uncertainty of missing values. (12) GP-VAE \cite{fortuin2020gp}: an architecture that combines the variational autoencoder and the Gaussian process to impute missing values. (13) rGAIN \cite{yoon2018gain}: a GAN-based method that employs a bidirectional recurrent encoder and decoder architecture. (14) CSDI \cite{tashiro2021csdi}: a model to impute multivariate time series with conditional score-based diffusion models, leveraging Transformer to extract the features. (15) SSSD \cite{alcaraz2022diffusion}: a diffusion-based method using structured state space models for time-series imputation. (16) PriSTI \cite{liu2023pristi}: a conditional diffusion framework for spatio-temporal data imputation with enhanced feature knowledge modeling. (17) PriSTI variant: PriSTI adopts \textit{our $4^{th}$-order conditional pseudo-numerical sampling method} instead of their first-order one from DDPM.

\renewcommand{\arraystretch}{1.1}
\begin{table*}[!t]
\centering
\caption{MAE and MSE/RMSE comparison with the baselines [\textbf{bold} = best, and \uline{underline} = second best]. (6) represents that our accelerated methods only requires 6 reverse steps during the inference.} \label{tab:Comparing}
\setlength{\tabcolsep}{7.5pt}
\begin{tabular}{ccccccccc}
\hline
                         & \multicolumn{4}{c}{\textbf{METR-LA}}                                                             & \multicolumn{4}{c}{\textbf{PEMS-BAY}}                                                           \\ \cline{2-9} 
Method                   & \multicolumn{2}{c}{Block-missing (16.52\%)} & \multicolumn{2}{c}{Point-missing (31.09\%)} & \multicolumn{2}{c}{Block-missing (9.20\%)} & \multicolumn{2}{c}{Point-missing (25.01\%)} \\ \cline{2-9} 
\textit{}                                      & MAE                 & MSE                  & MAE                 & MSE                  & MAE                 & MSE                 & MAE                 & MSE                  \\ \hline\hline
Mean                                           & 7.48±0.00           & 139.54±0.00          & 7.56±0.00           & 142.22±0.00          & 5.46±0.00           & 87.56±0.00          & 5.42±0.00           & 86.59±0.00           \\
KNN                  & 7.79±0.00           & 124.61±0.00          & 7.88±0.00           & 129.29±0.00          & 4.30±0.00           & 49.90±0.00          & 4.30±0.00           & 49.80±0.00           \\
Lin-ITP                                        & 3.26±0.00           & 33.76±0.00           & 2.43±0.00           & 14.75±0.00           & 1.54±0.00           & 14.14±0.00          & 0.76±0.00           & 1.74±0.00            \\
KF                                  & 16.75±0.00          & 534.69±0.00          & 16.66±0.00          & 529.96±0.00          & 5.64±0.00           & 93.19±0.00          & 5.68±0.00           & 93.32±0.00           \\
MICE             & 4.22±0.05           & 51.07±1.25           & 4.42±0.07           & 55.07±1.46           & 2.94±0.02           & 28.28±0.37          & 3.09±0.02           & 31.43±0.41           \\
VAR                                            & 3.11±0.08           & 28.00±0.76           & 2.69±0.00           & 21.10±0.02           & 2.09±0.10           & 16.06±0.73          & 1.30±0.00           & 6.52±0.01            \\ \hline
TRMF             & 2.96±0.00           & 22.65±0.13           & 2.86±0.00           & 20.39±0.02           & 1.95±0.01           & 11.21±0.06          & 1.85±0.00           & 10.03±0.00           \\
BATF                    & 3.56±0.01           & 35.39±0.03           & 3.58±0.01           & 36.05±0.02           & 2.05±0.00           & 14.48±0.01          & 2.05±0.00           & 14.90±0.06           \\
BRITS                      & 2.34±0.01           & 17.00±0.14           & 2.34±0.00           & 16.46±0.05           & 1.70±0.01           & 10.50±0.07          & 1.47±0.00           & 7.94±0.03            \\
GRIN                    & 2.03±0.00           & 13.26±0.05           & 1.91±0.00           & 10.41±0.03           & 1.14±0.01           & 6.60±0.10           & 0.67±0.00           & 1.55±0.01            \\ \hline
V-RIN            & 6.84±0.17           & 150.08±6.13          & 3.96±0.08           & 49.98±1.30           & 2.49±0.04           & 36.12±0.66          & 1.21±0.03           & 6.08±0.29            \\
GP-VAE                    & 6.55±0.09           & 122.33±2.05          & 6.57±0.10           & 127.26±3.97          & 2.86±0.15           & 26.80±2.10          & 3.41±0.23           & 38.95±4.16           \\
rGAIN                      & 2.90±0.01           & 21.67±0.15           & 2.83±0.01           & 20.03±0.09           & 2.18±0.01           & 13.96±0.20          & 1.88±0.02           & 10.37±0.20           \\
CSDI                    & 1.98±0.00           & 12.62±0.60           & 1.79±0.00           & 8.96±0.08            & 0.86±0.00           & 4.39±0.02           & 0.57±0.00           & 1.12±0.03            \\
SSSD               & 2.95±0.01           & 23.48±0.09           & 2.83±0.02           & 21.95±0.14           & 1.03±0.01           & 7.32±0.05           & 0.97±0.01           & 2.98±0.03            \\
PriSTI                    & 1.86±0.00           & 10.70±0.02           & \uline{1.72±0.00}           & 8.24±0.05            & 0.78±0.00           & 3.31±0.01           & 0.55±0.00           & 1.03±0.00            \\ 
PriSTI$^\dagger$ variant            & 1.83±0.00           & 10.57±0.00           & \uline{1.72±0.01}           & 8.20±0.00            & \uline{0.77±0.01}   & 3.29±0.01           & 0.52±0.00           & 1.01±0.01            \\ 
\hline
\textbf{FastSTI-2 (6)}    & \uline{1.81±0.01}  & \uline{10.44±0.00}    &  1.73±0.00                & \uline{8.17±0.03}      &  0.78±0.00       & \uline{3.28±0.03}   & \uline{0.51±0.00}           &  \uline{0.98±0.01}                           \\ 
\textbf{FastSTI-4 (6)}   & \textbf{1.79±0.01}  & \textbf{10.38±0.00}   &  \textbf{1.71±0.00}       & \textbf{8.15±0.02}     &  \textbf{0.75±0.00}       & \textbf{3.26±0.02}   & \textbf{0.50±0.00}   &  \textbf{0.96±0.01}                            \\ \hline
\hline
                         & \multicolumn{4}{c}{\textbf{PEMS04}}                                                             & \multicolumn{4}{c}{\textbf{PEMS08}}                                                           \\ \cline{2-9} 
                   & \multicolumn{2}{c}{Block-missing (10.59\%)} & \multicolumn{2}{c}{Point-missing (26.21\%)} & \multicolumn{2}{c}{Block-missing (9.41\%)} & \multicolumn{2}{c}{Point-missing (25.25\%)} \\ \cline{2-9} 
\textit{}      & MAE                 & RMSE                 & MAE                & RMSE                  & MAE                & RMSE                 & MAE                & RMSE     \\ \hline\hline
Mean           & 40.89±0.00          & 83.41±0.00           & 39.46±0.00         & 77.97±0.00            & 37.54±0.00         & 76.14±0.00          & 35.11±0.00         & 70.28±0.00     \\
KNN            & 39.28±0.00          & 75.41±0.00           & 35.14±0.00         & 66.90±0.00            & 33.49±0.00         & 70.25±0.00          & 31.06±0.00         & 63.41±0.00     \\
Lin-ITP       & 27.74±0.00           & 55.41±0.00           & 23.46±0.00         & 51.54±0.00            & 25.79±0.00         & 53.87±0.00          & 22.01±0.00         & 40.77±0.00     \\
KF            & 46.21±0.00           & 87.36±0.00          & 43.15±0.00          & 80.16±0.00            & 40.03±0.00         & 79.98±0.00          & 37.51±0.00         & 76.10±0.00     \\
MICE          & 29.56±0.01           & 57.84±0.04           & 27.15±0.06         & 50.28±0.36            & 27.94±0.01         & 49.14±0.09          & 24.76±0.02         & 39.97±0.07     \\
VAR           & 24.68±0.01           & 50.16±0.01           & 20.44±0.01         & 32.87±0.06            & 22.69±0.05         & 36.94±0.01          & 17.92±0.01         & 29.36±0.02     \\ \hline
TRMF          & 21.72±0.01           & 34.12±0.01           & 18.37±0.01         & 30.97±0.06            & 20.32±0.05         & 35.74±0.01          & 15.32±0.01         & 26.99±0.02     \\
BATF          & 27.10±0.13           & 40.28±0.01           & 20.89±0.01         & 34.11±0.02            & 23.87±0.05         & 37.88±0.03          & 18.26±0.03         & 29.21±0.02     \\
BRITS         & 20.24±0.01           & 33.05±0.02           & 17.87±0.03         & 29.74±0.01            & 18.37±0.02         & 32.51±0.01          & 14.67±0.01         & 25.33±0.03     \\
GRIN          & 18.77±0.03           & 29.88±0.08           & 16.94±0.01         & 26.11±0.01            & 12.98±0.02         & 24.23±0.10          & 11.37±0.01         & 18.21±0.01     \\ \hline
V-RIN         & 25.78±0.01           & 49.11±0.01          & 24.26±0.01          & 43.57±0.24            & 24.56±0.01         & 40.13±0.22          & 22.77±0.02         & 35.49±0.08     \\
GP-VAE        & 27.83±0.15           & 52.97±0.01          & 27.54±0.01          & 50.99±0.01            & 21.5±0.06          & 36.98±0.05          & 19.99±0.13         & 31.43±0.01     \\
rGAIN         & 17.67±0.01           & 29.84±0.07          & 15.21±0.04          & 25.03±0.01            & 14.03±0.01         & 24.72±0.01          & 10.98±0.02         & 19.34±0.01      \\
CSDI          & 16.58±0.01           & 28.13±0.02          & 15.09±0.01          & 24.91±0.02            & 12.06±0.05         & 23.08±0.10          & 10.07±0.01         & 17.19±0.00       \\
SSSD          & 17.93±0.03           & 30.19±0.01          & 15.32±0.04          & 25.11±0.01            & 13.96±0.01         & 24.63±0.01          & 10.97±0.01         & 19.21±0.01      \\
PriSTI        & 16.41±0.01           & 28.01±0.03          & 14.87±0.01          & 24.53±0.01            & 11.89±0.01         & 22.24±0.01          & 10.05±0.00         & 17.13±0.02     \\
PriSTI$^\dagger$ variant& 16.36±0.00           & 27.93±0.00          & 14.75±0.00          & 24.50±0.00            & 11.76±0.01         & 22.10±0.00          & 9.94±0.00          & 17.03±0.00     \\  \hline
\textbf{FastSTI-2 (6)}     & \uline{16.28±0.00}  & \uline{27.89±0.01}     &  \uline{14.69±0.01}  & \uline{24.47±0.03}      &  \textbf{11.60±0.01}      & \textbf{21.79±0.01}   & \uline{9.89±0.00} & \uline{16.97±0.01}                            \\ 
\textbf{FastSTI-4 (6)}     & \textbf{16.21±0.01}  & \textbf{27.70±0.00}     &  \textbf{14.67±0.02}  & \textbf{24.42±0.01}     &  \uline{11.73±0.01}       & \uline{22.00±0.00}   & \textbf{9.80±0.01}  & \textbf{16.93±0.01}                           \\ \hline\hline
\end{tabular}
\end{table*}

\renewcommand{\arraystretch}{1.1}
\begin{table*}[!t]
\centering
\caption{CPRS comparison ([\textbf{bold} = best, and \uline{underline} = second best]. \label{tab:CPRS} (6) represents that our accelerated methods only requires 6 reverse steps during the inference.}
\setlength{\tabcolsep}{2pt}
\begin{tabular}{ccccccccc}
\hline
\multirow{2}{*}{Method}  & \multicolumn{2}{c}{\textbf{METR-LA}}   & \multicolumn{2}{c}{\textbf{PEMS-BAY}}    & \multicolumn{2}{c}{\textbf{PEMS04}}    & \multicolumn{2}{c}{\textbf{PEMS08}}          \\ \cline{2-9} 
                         & Block-missing & Point-missing & Block-missing & Point-missing   & Block-missing & Point-missing & Block-missing & Point-missing       \\ \hline
                V-RIN    & 0.1283        & 0.7781        & 0.0394           & 0.0191       & 0.1842     &  0.1524          & 0.1522     &  0.1198                \\
                GP-VAE   & 0.1118        & 0.0977        & 0.0436           & 0.0568       & 0.1623     &  0.1358          & 0.1436     &  0.1103                \\
                CSDI     & 0.0260        & 0.0235        & 0.0127           & 0.0067       & 0.0583     &  0.0547          & 0.0420     &  0.0371                \\
                PriSTI   & 0.0244        & 0.0227        & \textbf{0.0093}  & 0.0064       & 0.0556     &  0.0522          & 0.0418     &  0.0364                \\ \hline
\textbf{FastSTI-2 (6)}       & \uline{0.0243}   & \uline{0.0223}  & \uline{0.0095}   & \uline{0.0062}   & \uline{0.0542}    & \uline{0.0505}    & \textbf{0.0373}    & \textbf{0.0307} \\ 
\textbf{FastSTI-4 (6)}       & \textbf{0.0241}  & \textbf{0.0219} & \textbf{0.0093}  & \textbf{0.0060}  & \textbf{0.0540}   & \textbf{0.0503}   & \uline{0.0374}     & \textbf{0.0305}  \\ \hline
\end{tabular}
\end{table*}

\subsection{Imputation Performance} \label{sec:8}

First, we compare our model to the seventeen baselines. Table \ref{tab:Comparing} reports the MAE and MSE/RMSE comparison on two traffic speed datasets (METR-LA and PEMS-BAY) and two traffic flow datasets (PEMS04 and PEMS08), while Table \ref{tab:CPRS} reports the CRPS metric. FastSTI, using only 6 reverse steps, outperforms all of the baselines, producing more realistic imputation. Focusing on the comparison between FastSTI and PriSTI \cite{liu2023pristi}, we have a better performance, proving the effectiveness of the proposed high-order conditional pseudo-numerical method in generating higher-quality samples. Notably, traffic flow is more challenging than traffic speed because its fluctuations are more significant and random over time. Our approach enhances overall performance on traffic flow datasets (PEMS04 and PEMS08) by utilizing higher-order pseudo-numerical methods that better handle noise and uncertainty in the data. Conversely, PriSTI\cite{liu2023pristi}, CSDI\cite{tashiro2021csdi}, and SSSD\cite{alcaraz2022diffusion} exhibit limited effectiveness, attributed to their first-order numerical sampling from DDPM \cite{ho2020denoising}, which struggles with the complexities of traffic patterns. FastSTI also benefits from the Diff-GCN features extractor, in contrast to the message-passing neural network (MPNN) used in the best baseline PriSTI. Meanwhile, we applied the proposed sampling method on PriSTI, called the PriSTI variant, and the results show that it can help improve the interpolation performance. Additionally, the comparison between FastSTI-2 and FastSTI-4 favors FastSTI-4, which makes sense, as the imputation quality can improve as the order of the numerical method increases.

\begin{figure}[t]
    \centering
    \subfloat[\scriptsize{Block-missing scenario}]{\includegraphics[width=.5\columnwidth]{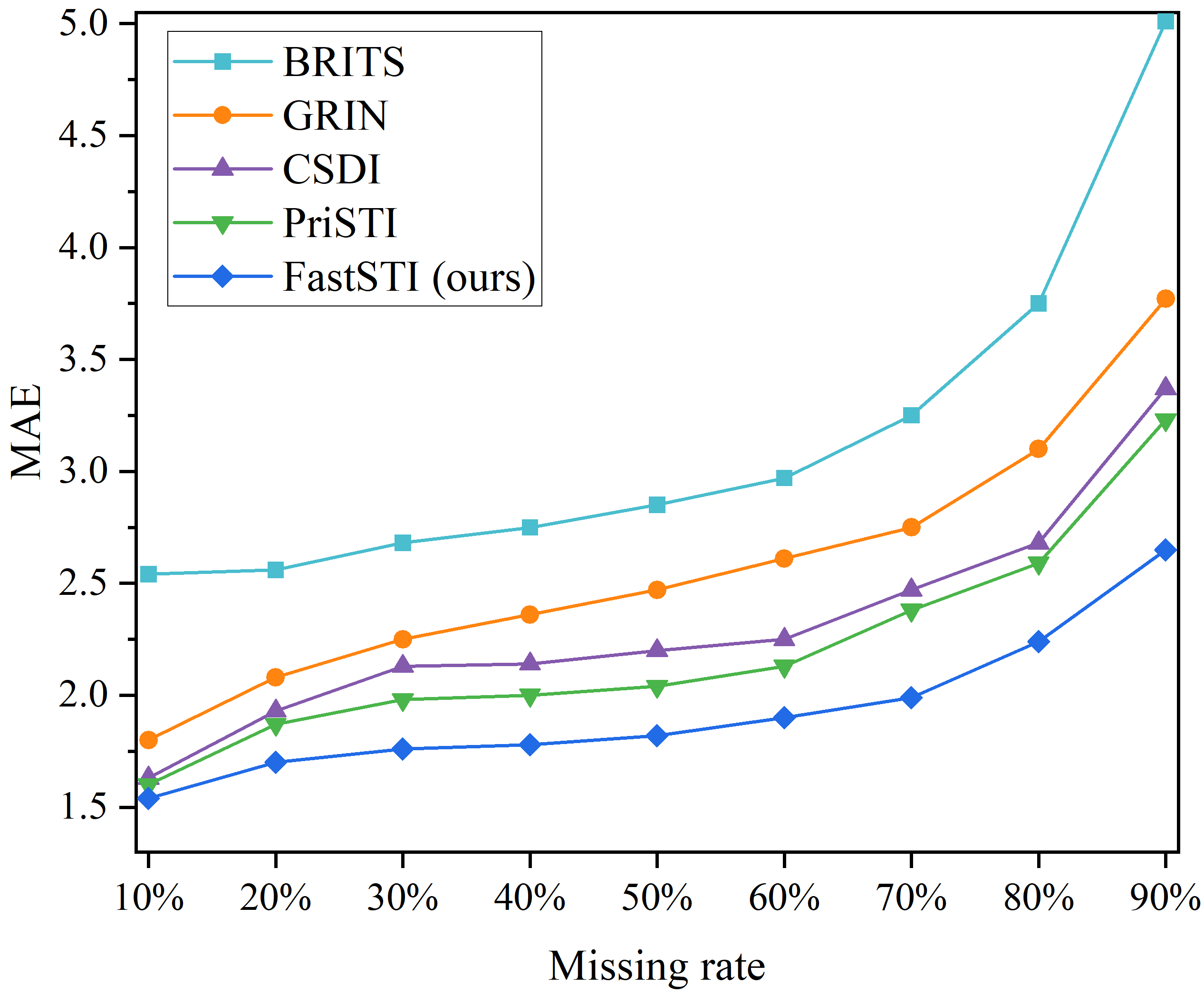}%
    \label{fig:Block-missing on METR-LA}}
    \hfil
    \subfloat[\scriptsize{Point-missing scenario}]{\includegraphics[width=.5\columnwidth]{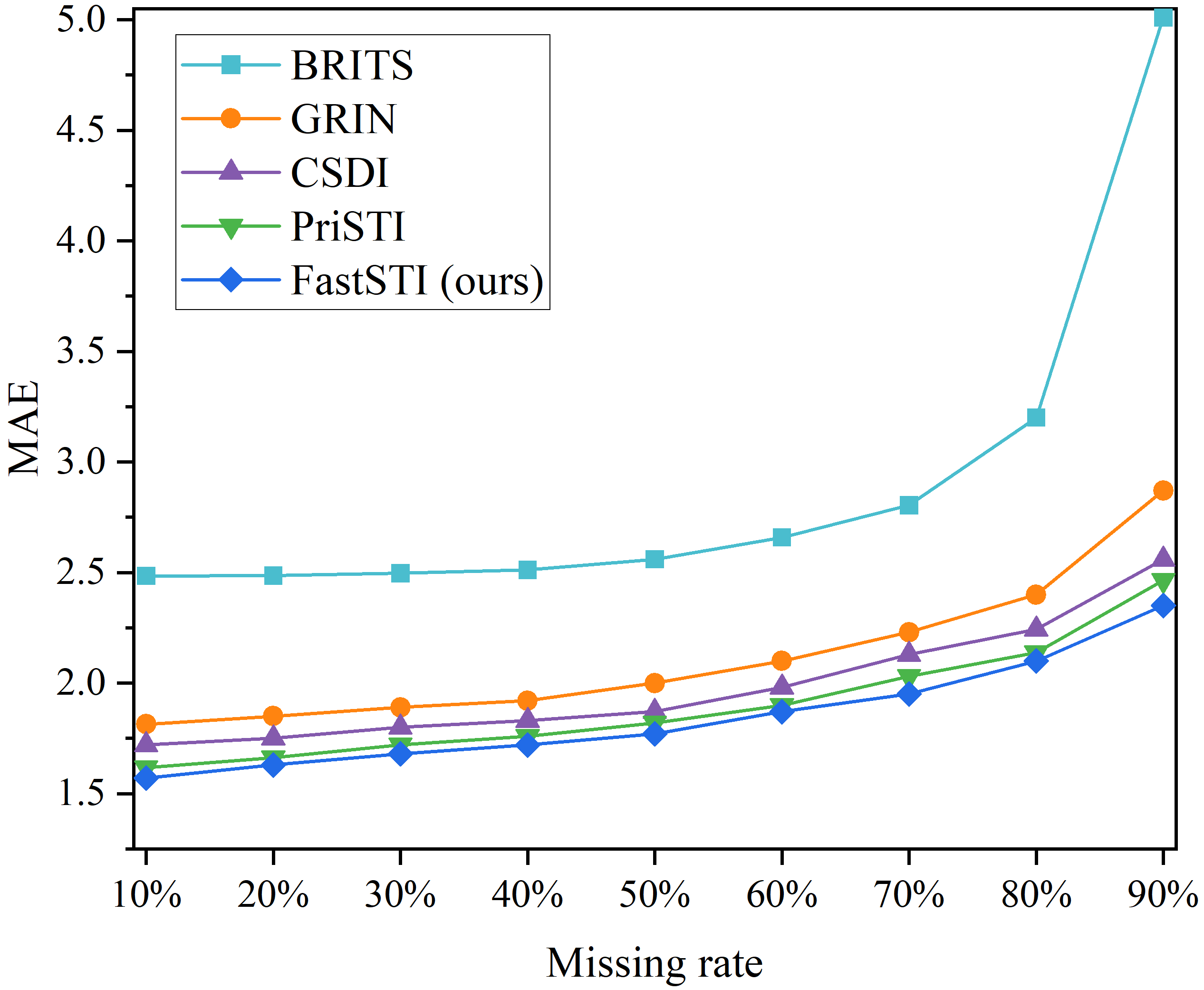}%
    \label{fig:Point-missing on METR-LA}}
    \caption{The Impact of Missing Rate on the Imputation Performance on METR-LA Dataset.}
    \label{fig:The imputation performance of block-missing and point-missing rates}
\end{figure}

\begin{figure}[t]
    \centering
    \subfloat[\scriptsize{Block-missing scenario}]{\includegraphics[width=.5\columnwidth]{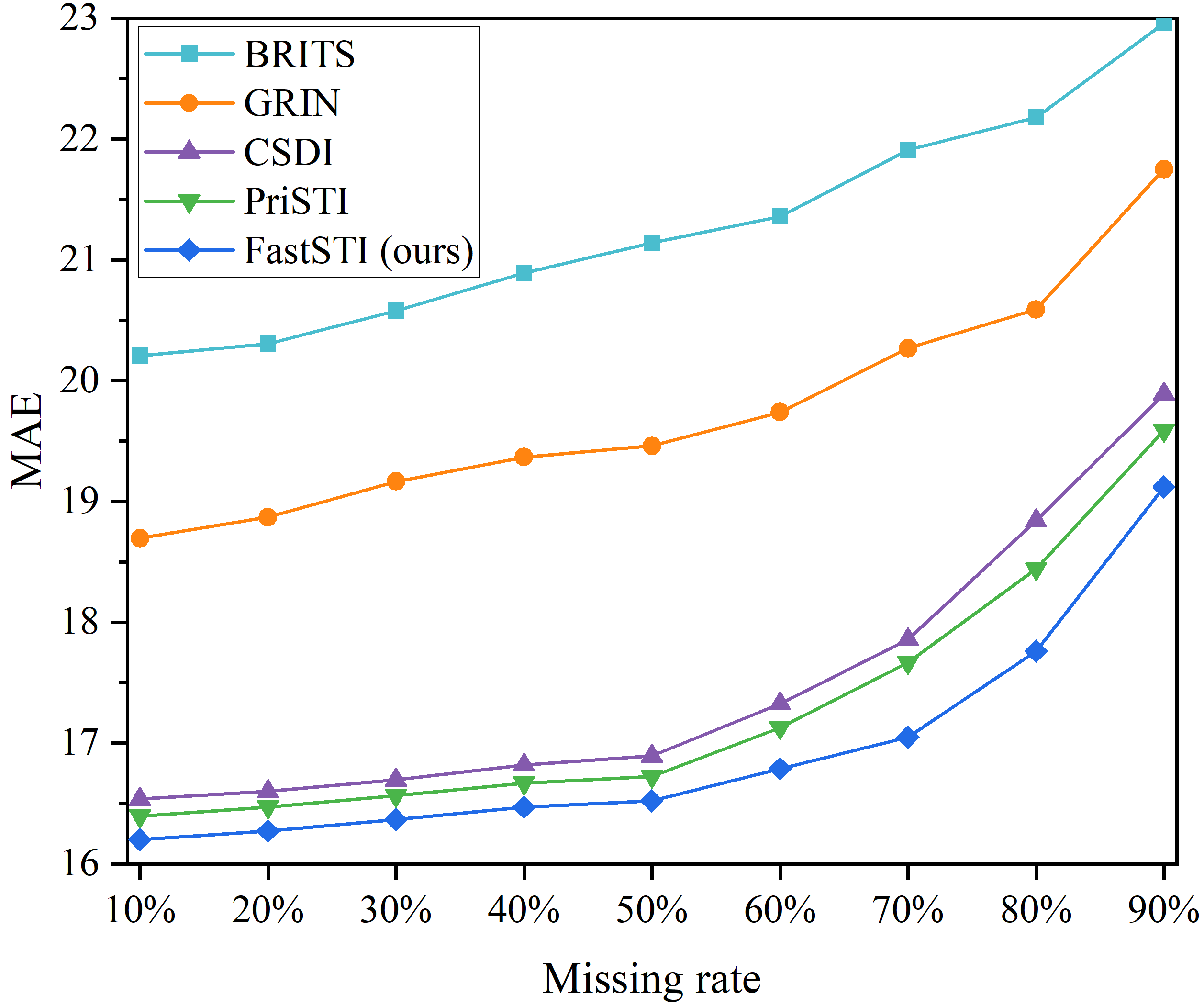}%
    \label{fig:Block-missing on PEMS04}}
    \hfil
    \subfloat[\scriptsize{Point-missing scenario}]{\includegraphics[width=.5\columnwidth]{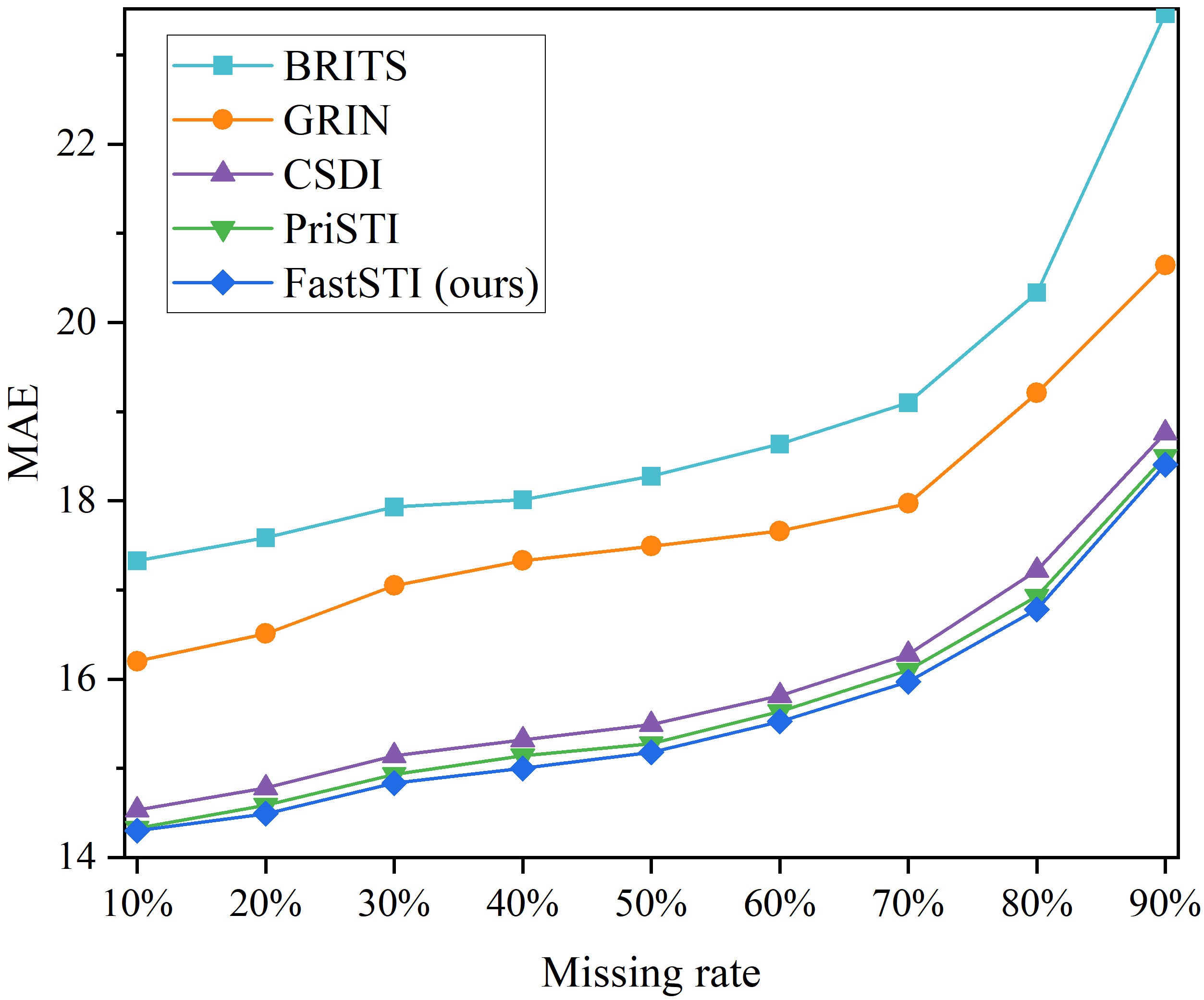}%
    \label{fig:Point-missing on PEMS04}}
    \caption{The Impact of Missing Rate on the Imputation Performance on PEMS04 Dataset.}
    \label{fig:The imputation performance of block-missing and point-missing rates PEMS04}
\end{figure}

To comprehensively evaluate the imputation performance under different missing rates, we report missing rates from 10\% to 90\% on the METR-LA and PEMS04 datasets in Figures \ref{fig:The imputation performance of block-missing and point-missing rates} and \ref{fig:The imputation performance of block-missing and point-missing rates PEMS04}. Figure \ref{fig:Block-missing on METR-LA} and \ref{fig:Block-missing on PEMS04} show such an effect with the block-missing strategy, and Figure \ref{fig:Point-missing on METR-LA} and \ref{fig:Point-missing on PEMS04} show the impact of increasing the point-missing rate. Intuitively, as the rate of missing values increases, data imputation becomes more challenging due to the reduced availability of observed values and the increased complexity of extracting spatiotemporal correlation of traffic conditions. We can see that the proposed FastSTI-4 (6) consistently outperforms baseline models in terms of imputation performance, regardless of the missing rate. 1) As show in Figures \ref{fig:Block-missing on METR-LA} and \ref{fig:Point-missing on METR-LA}, FastSTI outperforms the competitors by up to 47.11\% in MAE (FastSTI-4 vs. BRITS block-missing at 90\% rate) and 42.4\% in MAE (FastSTI-4 vs. BRITS point-missing at 90\% rate). Compared with the best baseline model PriSTI, FastSTI-4 reduces the MAE by up to 17.96\% (block-missing at 90\% rate) and 7.3\% (point-missing at 90\% rate). 2) From a block-missing rate of 10\% to 90\% (see Fig. \ref{fig:Block-missing on PEMS04}), the error of the FastSTI increased by 18.01\%, while that of the PriSTI increased by 19.47\%. FastSTI is less influenced by changes in the missing rate and has demonstrated better robustness than PriSTI. These results also prove the effectiveness of FastSTI in keeping better accurate imputation, especially under high missing rates (60\% $\sim$ 90\%), by effectively capturing the spatial and temporal correlations among traffic speed observations across various locations and periods. Moreover, compared to baseline models, our advanced conditional pseudo-numerical method can extract additional features from observed values as the missing rate increases.

\subsection{Time Performance} \label{sec:9}

We evaluate the inference time of our proposed FastSTI from two perspectives: 1) with and without our acceleration technique, and 2) with two different random walk steps $K \in \{2,6\}$. The detailed settings are as follows.

1) \textit{Without acceleration}:
\begin{itemize}
    \item \textbf{CSDI \cite{tashiro2021csdi}}. CSDI with 50 reverse steps, following the original paper settings.
    \item \textbf{PriSTI \cite{liu2023pristi}}. PriSTI with 50 reverse steps, following the original paper settings.
    \item \textbf{FastSTI-2 ($K$=6) w/o acc}. FastSTI-2 with 50 reverse steps and $K$=6 of Diff-GCN.
    \item \textbf{FastSTI-4 ($K$=6) w/o acc}. FastSTI-4 with 50 reverse steps and $K$=6 of Diff-GCN.
    \item \textbf{FastSTI-2 ($K$=2) w/o acc}. FastSTI-2 with 50 reverse steps and $K$=2 of Diff-GCN.
    \item \textbf{FastSTI-4 ($K$=2) w/o acc}. FastSTI-4 with 50 reverse steps and $K$=2 of Diff-GCN.
\end{itemize}

2) \textit{Acceleration}:
\begin{itemize}
    \item \textbf{PriSTI w/ our acc}. PriSTI of with 6 reverse steps, integrated with \textit{our acceleration method} (see Section \ref{sec:10}).
    \item \textbf{FastSTI-2 ($K$=6) w/ acc}. FastSTI-2 with 6 reverse steps and $K$=6 of Diff-GCN.
    \item \textbf{FastSTI-4 ($K$=6) w/ acc}. FastSTI-4 with 6 reverse steps and $K$=6 of Diff-GCN.
    \item \textbf{FastSTI-2 ($K$=2) w/ acc}. FastSTI-2 with 6 reverse steps and $K$=2 of Diff-GCN.
    \item \textbf{FastSTI-4 ($K$=2) w/ acc}. FastSTI-4 with 6 reverse steps and $K$=2 of Diff-GCN.
\end{itemize}

\begin{figure}[t]
    \centering
    \includegraphics[width=.83\linewidth]{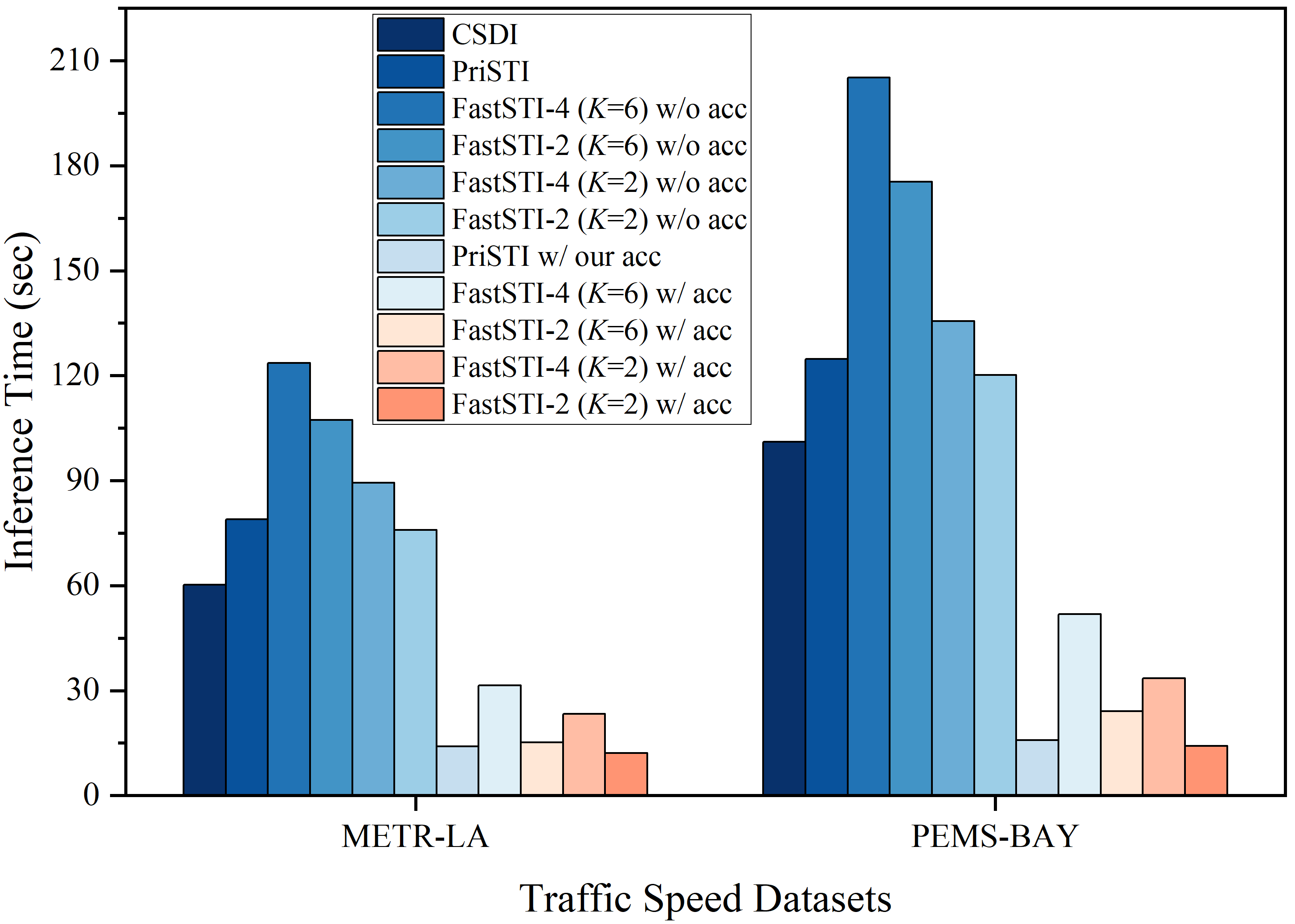}
    \caption{Inference Times on Traffic Speed Datasets (METR-LA and PEMS-BAY).}
    \label{fig:The time cost of FastSTI Speed}
\end{figure}

\begin{figure}[t]
    \centering
    \includegraphics[width=.83\linewidth]{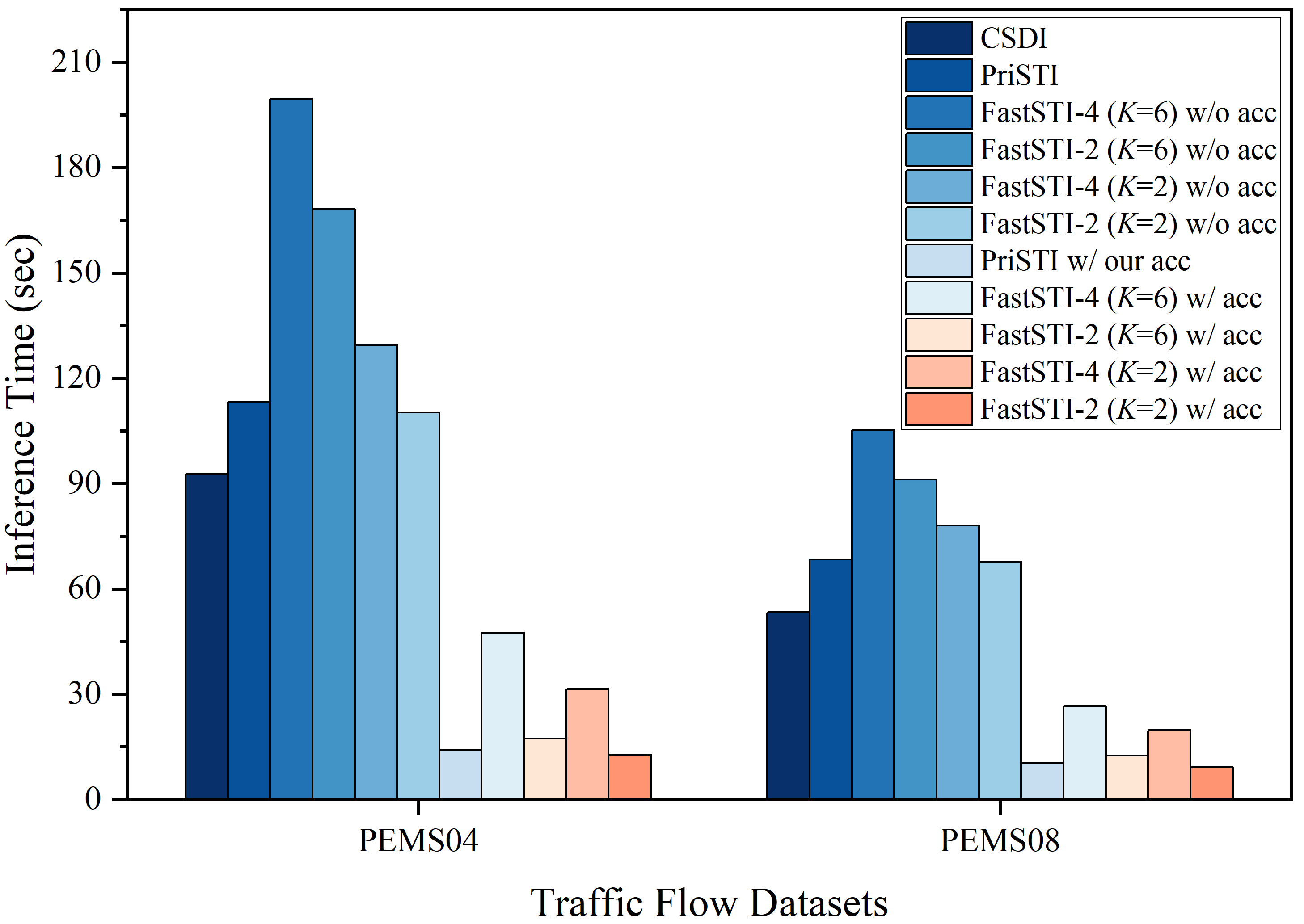}
    \caption{Inference Times on Traffic Flow Datasets (PEMS04 and PEMS08).}
    \label{fig:The time cost of FastSTI Flow}
\end{figure}

Figures \ref{fig:The time cost of FastSTI Speed} and \ref{fig:The time cost of FastSTI Flow} display the imputation time of the models for all sensors throughout 24-time points (i.e., for 2 hours at 5-minute intervals) on traffic speed and flow datasets. We can conclude that: 1) FastSTI utilizes a tuned variance schedule to speed up inference time in the imputation phase, which enables FastSTI to impute highly accurate traffic data with only 6 reverse steps ($8.3 \times$ less than the $50$ steps of both PriSTI \cite{liu2023pristi} and CSDI \cite{tashiro2021csdi}). Meanwhile, from Figure \ref{fig:The time cost of FastSTI Speed}, our FastSTI-2 ($K$=2) w/ acc. ($\sim 12.10s$) is $~6.5 \times$ faster than PriSTI ($\sim 78.96s$) and $~5 \times$ faster than CSDI ($\sim 60.2s$) on METR-LA. With the higher-order FastSTI-4 w/ acc. ($\sim 31.44s$), the model requires more inference time; however, FastSTI-4 ($K$=2) w/ acc is still $~3.4 \times$ faster than PriSTI and $~2.5$x faster than CSDI. Thereby, FastSTI significantly reduces the computational time required compared to these competing diffusion architectures. 2) The larger the parameters of Diff-GCN, the slower the inference speed. As shown in Figure 6, with the accelerated version, FastSTI-2 ($K$=2) is 18.7 seconds faster than FastSTI-2 ($K$=6) at every two hours of imputation on the PEMS04 dataset. Similarly, FastSTI-2 ($K$=2) is 10.6 seconds faster than FastSTI-2 ($K$=6) on PEMS08. This indicates that increasing the $K$ parameter in Diff-GCN allows for more geospatial information extraction from traffic patterns, but it also adds to the computational time burden. 3) Our acceleration technique can be compatible with most diffusion-based models (e.g., PriSTI \cite{liu2023pristi}). Compared to our FastSTI-2 ($K$=2), PriSTI can obtain a close inference time using our acceleration method.

\begin{table}[!t]
\centering
\caption{The influence of different components on FastSTI-4. \label{tab:Ablation study}}
\setlength{\tabcolsep}{5pt}
\begin{tabular}{cccccc}
\hline
\multirow{3}{*}{PN} &   \multirow{3}{*}{Diff-GCN}    & \multicolumn{4}{c}{\textbf{METR-LA}}                                           \\ \cline{3-6} 
                        &     & \multicolumn{2}{c}{Block-missing} & \multicolumn{2}{c}{Point-missing} \\ \cline{3-6} 
                         &    & MAE                & MSE                   & MAE                      & MSE                   \\ \hline
\xmark & \xmark & 1.86±0.00 & 10.70±0.02 & 1.72±0.00 & 8.24±0.05\\
\xmark & \cmark                       & 1.83±0.00          & 10.40±0.00            &  1.74±0.00               & 8.22±0.00             \\
\cmark & \xmark                 & 1.89±0.00          & 10.87±0.00            &  1.82±0.00               & 8.57±0.00             \\
\hline
\cmark & \cmark & \textbf{1.79±0.01} & \textbf{10.38±0.00}   &  \textbf{1.71±0.00}      & \textbf{8.15±0.02}    \\ \hline
\end{tabular}
\end{table}

\begin{table*}[]
\caption{The impact of increasing the number of steps on FastSTI \label{tab:num_steps}. (6) represents that FastSTI only requires 6 reverse steps \textit{WITH} our accelerated method, and (50) represents FastSTI use 50 reverse steps \textit{WITHOUT} accelerated method.}
\resizebox{\textwidth}{!}{%
\begin{tabular}{cccc|ccc|ccc|ccc}
\hline
                         & \multicolumn{6}{c}{\textbf{METR-LA}}                                                             & \multicolumn{6}{c}{\textbf{PEMS-BAY}}                                                           \\ \cline{2-13} 
Method                   & \multicolumn{3}{c}{Block-missing (16.52\%)} & \multicolumn{3}{c}{Point-missing (31.09\%)} & \multicolumn{3}{c}{Block-missing (9.20\%)} & \multicolumn{3}{c}{Point-missing (25.01\%)} \\ \cline{2-13}
& MAE & MSE & CRPS & MAE & MSE & CRPS & MAE & MSE & CRPS & MAE & MSE & CRPS \\ \hline

\textbf{FastSTI-2 (50)}                  & 1.80±0.00   & 10.34±0.01 & 0.0241        & 1.69±0.03         & 8.01±0.03   & {0.0220} & 0.72±0.01           & 3.26±0.00 & 0.0091          & 0.52±0.01           & 1.01±0.02    & 0.0061                         \\ 
\textbf{FastSTI-4 (50)}                  & \textbf{1.79±0.01}  & \textbf{10.29±0.00} & \textbf{0.0239}          & \textbf{1.68±0.00}    & \textbf{7.99±0.01} & \textbf{0.0219}  & \textbf{0.70±0.00}          & \textbf{3.24±0.00} & \textbf{0.0089}           & 0.51±0.00           & 0.99±0.00      & \textbf{0.0059}                       \\ \hline
\textbf{FastSTI-2 (6)}                   & 1.81±0.01           & 10.44±0.00            & 0.0243    &  1.73±0.00                & 8.17±0.03 & 0.0223           &  0.78±0.00       & 3.28±0.03  & 0.0095  & 0.51±0.00           &  0.98±0.01 & 0.0062                          \\ 
\textbf{FastSTI-4 (6)}                   & \textbf{1.79±0.01}  & 10.38±0.00 & 0.0241               &  1.71±0.00                & 8.15±0.02   & \textbf{0.0219}        &  0.75±0.00       & 3.26±0.02 & 0.0093  & \textbf{0.50±0.00}   &  \textbf{0.96±0.01} & 0.0060     \\ \hline
\hline
                         & \multicolumn{6}{c}{\textbf{PEMS04}}                                                             & \multicolumn{6}{c}{\textbf{PEMS08}}                                                           \\ \cline{2-13} 
Method                   & \multicolumn{3}{c}{Block-missing (10.59\%)} & \multicolumn{3}{c}{Point-missing (26.21\%)} & \multicolumn{3}{c}{Block-missing (9.41\%)} & \multicolumn{3}{c}{Point-missing (25.25\%)} \\ \cline{2-13}
& MAE & RMSE & CRPS & MAE & RMSE & CRPS & MAE & RMSE & CRPS & MAE & RMSE & CRPS \\ \hline

\textbf{FastSTI-2 (50)}    & 16.22±0.00   & 27.84±0.01 & 0.0539   & 14.64±0.01  & 24.42±0.03  & 0.0503   & \textbf{11.57±0.01}  & \textbf{21.77±0.02} & \textbf{0.0371}   & 9.82±0.02  & 16.93±0.01  & \textbf{0.0302}                         \\ 
\textbf{FastSTI-4 (50)}    & \textbf{16.17±0.01}  & \textbf{27.64±0.03} & \textbf{0.0537}       & \textbf{14.63±0.01}   & \textbf{24.38±0.01} & \textbf{0.0501}  & 11.69±0.01         & 21.95±0.02 & 0.0373           & \textbf{9.74±0.01} & \textbf{16.89±0.01} & \textbf{0.0302}                       \\ \hline
\textbf{FastSTI-2 (6)}     & 16.28±0.00  & 27.89±0.01  & 0.0542    &  14.69±0.01  & 24.47±0.03 & 0.0505           &  11.60±0.01      & 21.79±0.01  & 0.0373  & 9.89±0.00 & 16.97±0.01   & 0.0307                          \\ 
\textbf{FastSTI-4 (6)}     & 16.21±0.01  & 27.70±0.00 & 0.0540     &  14.67±0.02  & 24.42±0.01   & 0.0503        &  11.73±0.01       & 22.00±0.00 & 0.0374  & 9.80±0.01  & 16.93±0.01 & 0.0305     \\ \hline\hline
\end{tabular}}
\end{table*}

\begin{table}[t]
\centering
\caption{Influence of different diffusion parameters on METR-LA dataset [\textbf{bold} = best]. \label{tab:Diffusion Parameters}}
\setlength{\tabcolsep}{13pt}
\begin{tabular}{ccc|cc}
\hline
\multicolumn{3}{{c}}{Diffsion parameters}                    & \multicolumn{2}{c}{Performance} \\ \hline
$\beta _{1}$            & $\beta _{T} $        & schedule  & MAE            & MSE            \\ \hline
\multirow{3}{*}{0.0001} & \multirow{3}{*}{0.2} & linear    & 1.96           & 12.46          \\ 
                        &                      & cosine    & 1.88           & 11.78          \\
                        &                      & quadratic & \textbf{1.79}  & \textbf{10.38} \\ \hline
0.001                   & 0.2                  & quadratic & 2.06           & 13.87          \\ \hline
\multirow{3}{*}{0.0001} & 0.1                  & linear    & 1.95           & 12.37          \\
                        & 0.1                  & quadratic & 1.86           & 11.54          \\
                        & 0.3                  & quadratic & 1.91           & 11.89          \\ \hline
\end{tabular}
\end{table}

\subsection{Ablation Study on Imputation Performance}
We conduct an ablation study on the METR-LA datasets to verify the impact of different components of our FastSTI model. Table \ref{tab:Ablation study} illustrates the performance of FastSTI-4 with and without employing our proposed components, including the pseudo-numerical method (PN) and the GCN-based conditional extractor (Diff-GCN). When we remove the pseudo-numerical (PN) methods, the model no longer considers high-order numerical methods converging to the exact solution when $\Delta t$ is closer to 0. Consequently, it no longer utilizes a larger iteration interval $\Delta t$ to achieve global error reduction, resulting in decreased imputation accuracy. Additionally, the spatial learning sub-component, Diff-GCN, is crucial in extracting geographic interactions among nodes within the spatial correlation. When removing Diff-GCN, the model has a weak ability to capture the influence among nodes by spreading the traffic feature information on graph $G$. Therefore, integrating PN and Diff-GCN together into the model achieves the best performance.

For further ablation, we investigate the impact of increasing the number of denoising steps on our model. Table \ref{tab:num_steps} reports a comparison between our FastSTI (6 steps) and FastSTI \textit{w/o acceleration} when increasing the number of steps to 50. Surly, increasing the number of steps increases the performance by a noticeable gap, proving the effectiveness of FastSTI quality-wise. However, increasing the number of steps results in a dramatic increase in imputation time.

\begin{figure}[t]
    \centering
    \subfloat[\scriptsize{PEMS04 Block-missing}]{\includegraphics[width=.4\columnwidth]{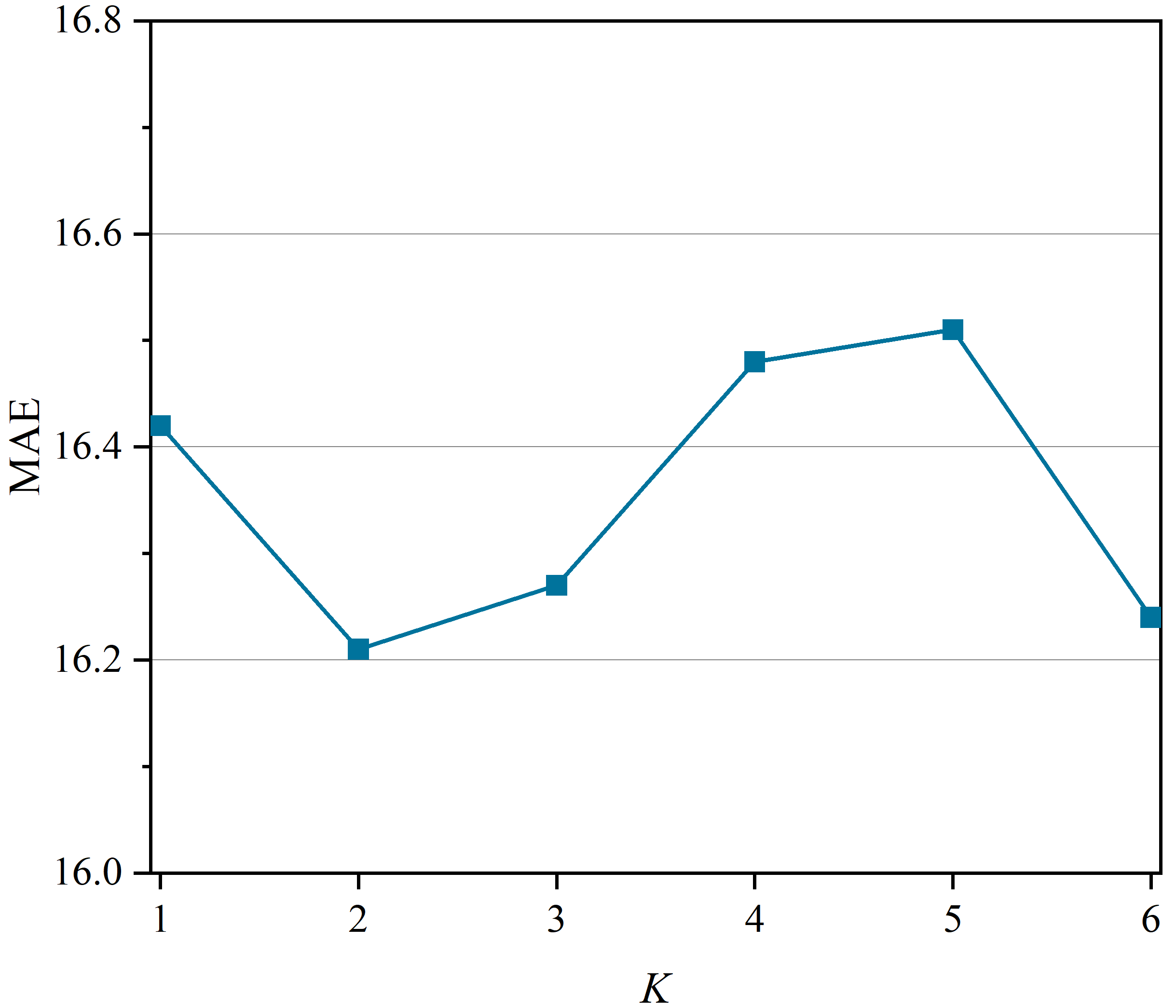}%
    \label{fig:PEMS04-K}}
    \hfil
    \subfloat[\scriptsize{PEMS08 Block-missing}]{\includegraphics[width=.4\columnwidth]{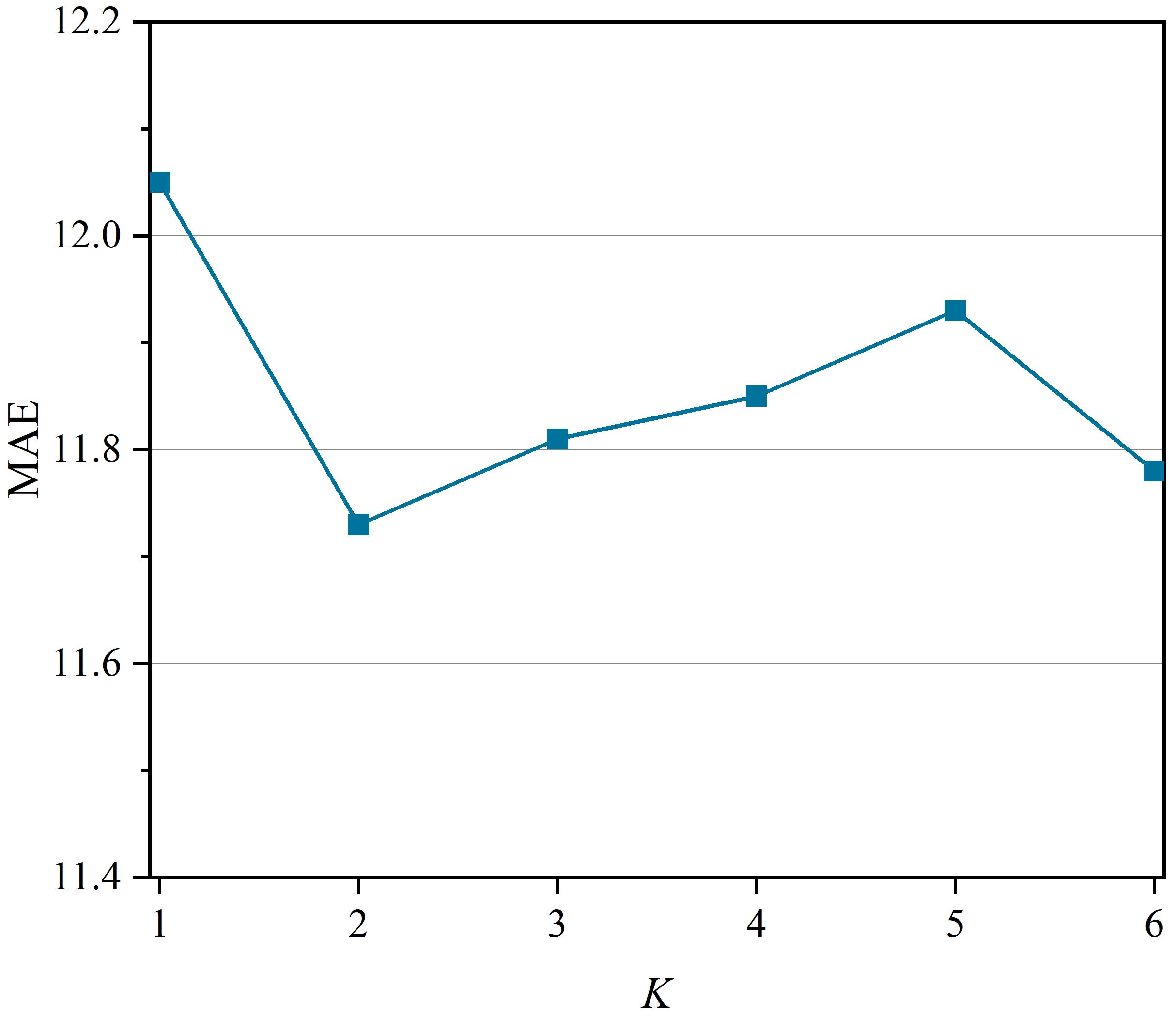}%
    \label{fig:PEMS08-K}}
    \caption{The Effect of Diff-GCN's $K$ Parameter on Imputation Performance.}
    \label{fig:The imputation performance of K parameter}
\end{figure}
\begin{figure}[t]
    \centering
    \subfloat[\scriptsize{PEMS04 Block-missing}]{\includegraphics[width=.4\columnwidth]{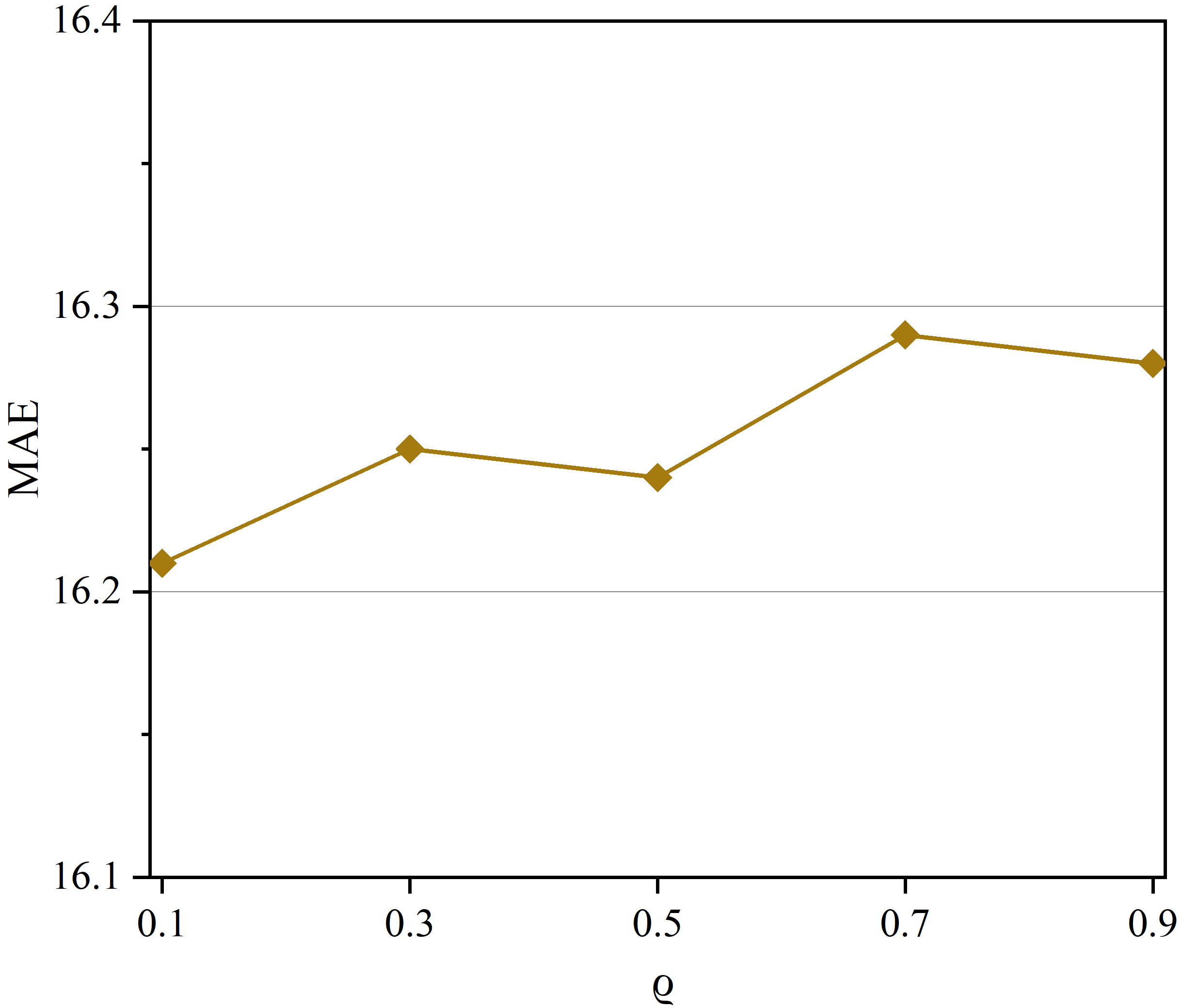}%
    \label{fig:PEMS04_varrho}}
    \hfil
    \subfloat[\scriptsize{PEMS08 Block-missing}]{\includegraphics[width=.4\columnwidth]{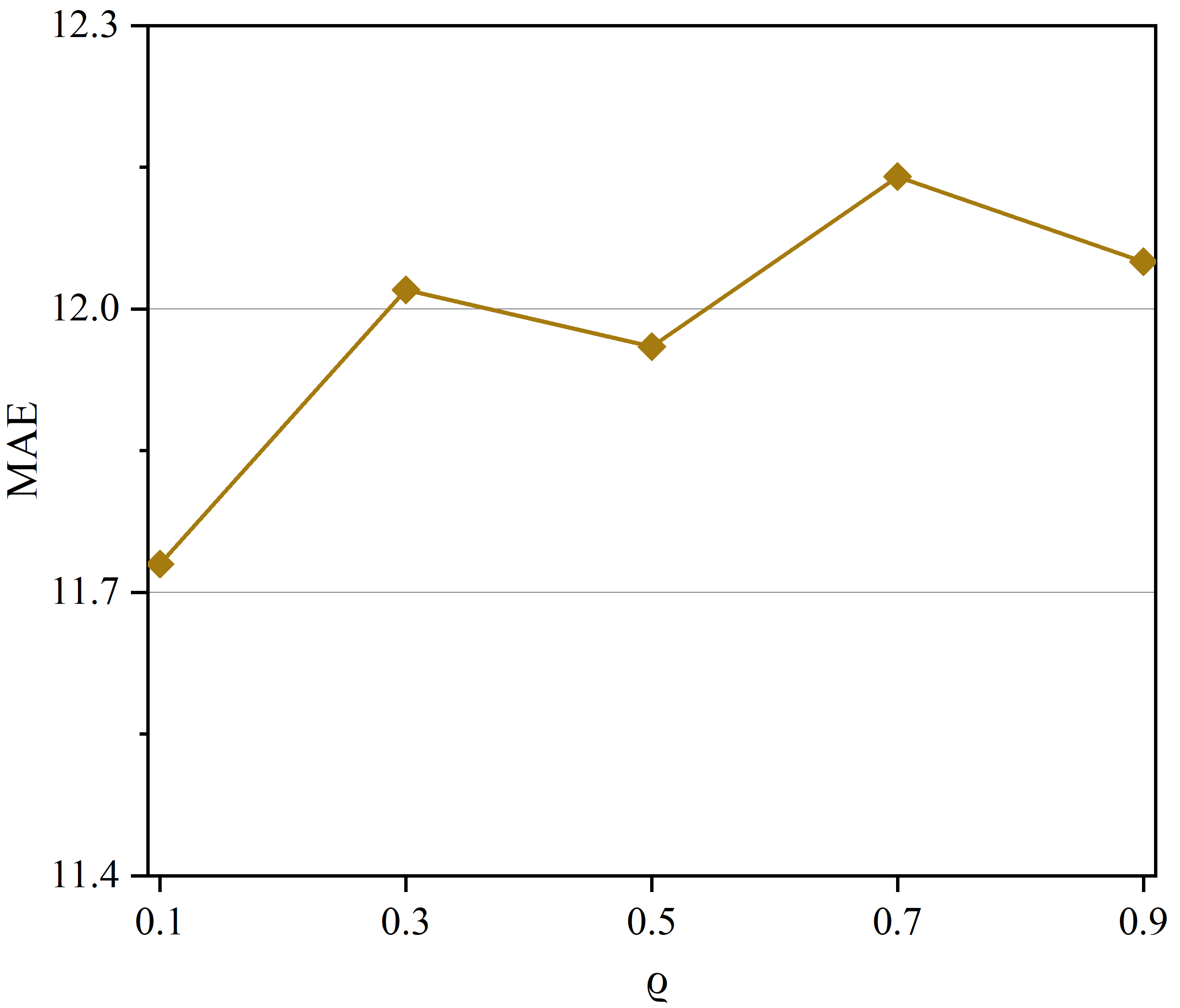}%
    \label{fig:PEMS08_varrho}}
    \caption{The Effect of Diff-GCN's $\varrho$ Parameter on Imputation Performance.}
    \label{fig:The imputation performance of varrho parameter}
\end{figure}

\subsection{Hyperparameter Analysis}
In this part, we explore the impact of FastSTI's diffusion parameters on the METR-LA dataset, including the minimum noise level $\beta _{1}$, the maximum noise level $\beta _{T} $, and the diffusion schedule to generate $\beta$. Table \ref{tab:Diffusion Parameters} reveals that FastSTI performs best when setting $\beta_1$ to 0.0001, $\beta_T$ to 0.2, and utilizing a quadratic diffusion schedule. Here, the noise level parameter is employed to regulate the diffusion speed. The convergence speed of the model slows down when the $\beta _{T} $ is smaller, while the numerical stability of the model is compromised when the $\beta _{T} $ is larger. Additionally, regarding the diffusion schedule, a quadratic schedule outperforms linear and cosine schedules. This is because the quadratic schedule allows for a gentle decay of $\alpha_t$, which in turn improves sample quality, making it the optimal choice for FastSTI.

To further investigate our Diff-GCN feature extractor for capturing local geospatial dependencies, we report different parameters of graph random walk step $K$ and the graph coefficient $\varrho$ of FastSTI-4 on the PEMS04 and PEMS08 datasets under the block-missing scenario. 
1) The parameter $K$ corresponds to the convolutional filter's receptive field size. Large $K$ values enable the module to capture a wide range of spatial graph information but come with the trade-off of increased learning complexity. As shown in Figures \ref{fig:PEMS04-K} and \ref{fig:PEMS08-K}, we can observe that with increasing the number of $K$, the error on the testing set initially decreases, followed by a slight increase. Compared with $K$=2, $K$=6 reports similar MAE values but heavily increases the time consumption. We also analyze the time performance of the setting of these two parameters in Sec. \ref{sec:9}. Therefore, $K$=2 can be selected as the optimal parameter. That is, the graph convolution operation can aggregate the features of the node itself, as well as those of its $1$-hop and $2$-hop graph neighbors when $K$=2. 
2) The graph coefficient $\varrho$ represents the scaling factor controlling the influence of the neighboring nodes' matrix (excluding self-node) during feature propagation in the Diff-GCN. For road traffic pattern, where the self-node spatial correlation typically holds greater significance than neighboring nodes' contributions, setting the optimal $\varrho$=0.1 (see Fig. \ref{fig:The imputation performance of varrho parameter}) means that the influence of $K$-hop (where $K>0$) node's adjacency matrix on the feature propagation is scaled to 10\% of their original value. Such configuration enables the model to primarily focus on a self-node adjacency matrix for feature propagation while slightly considering the impact of the neighboring nodes.

\subsection{Case Study}
\begin{figure*}[!ht]
    \centering
    \includegraphics[width=.75\linewidth]{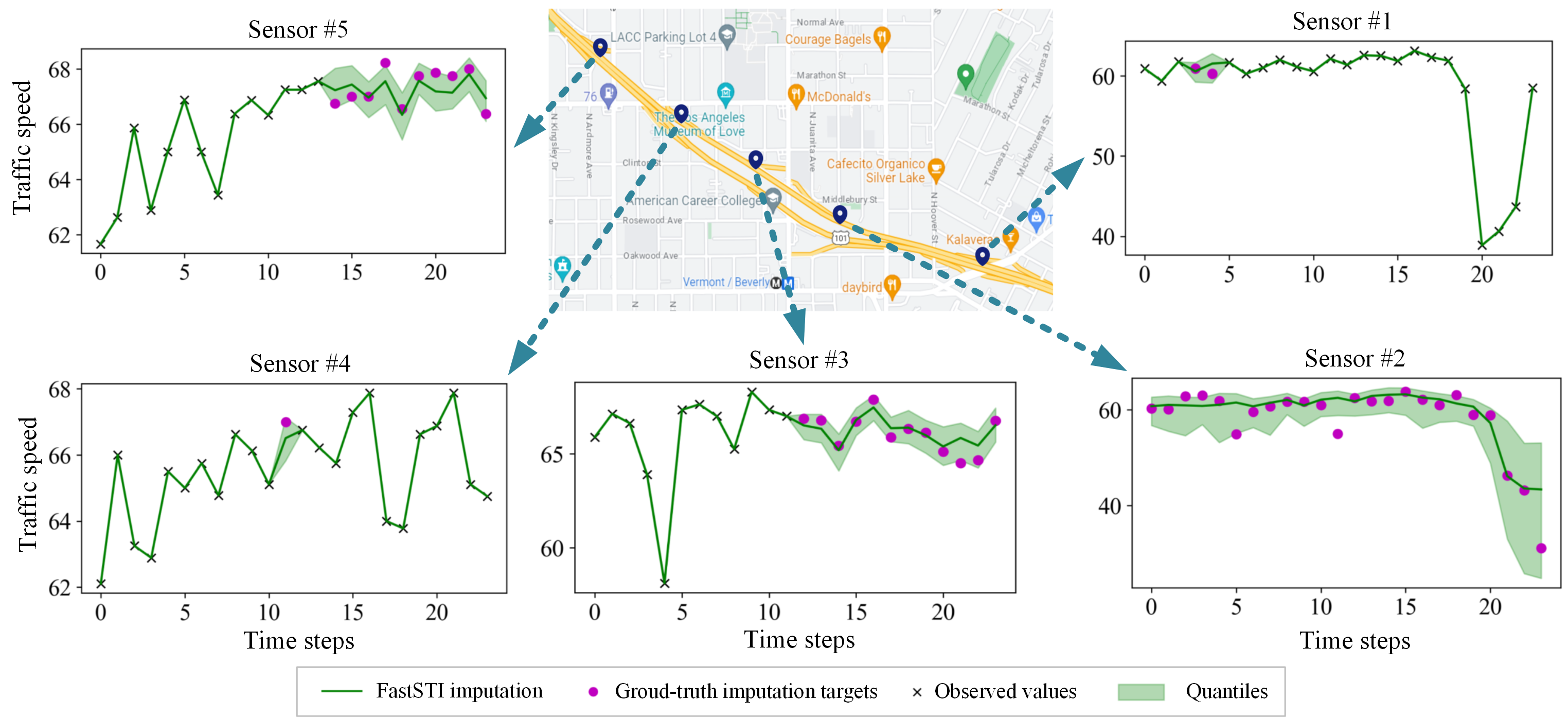}
    \caption{Case study on METR-LA.}
    \label{fig:The case Study on METR-LA}
\end{figure*} 
We present the visualized results of the case study on the METR-LA dataset; comparing METR-LA to PEMS-BAY (Bay Area) shows that traffic imputation on the METR-LA dataset (Los Angeles, famous for its complicated traffic conditions) is a more complicated way. Taking the block-missing scenario as an example, Fig. \ref{fig:The case Study on METR-LA} provides a visualization of the two-hour imputation (24-time points) for five closely located sensors/nodes. Here, the green lines represent the imputed values generated by our FastSTI-4. The purple points indicate the ground-truth imputation targets. The black crosses represent the observed values, while the green shade is the quantiles between 5\% and 95\%. As the figure shows, sensors \#1 and \#4 exhibit minimal instances of missing data, while sensors \#3 and \#5 show a continuous pattern of missing values spanning approximately an hour. Moreover, in more extreme cases, i.e., sensor \#2, there is a complete absence of usable observed value for two hours. It is also apparent that our FastSTI-4 shows positive results in most cases. For instance, sensor \#5 can utilize its temporal feature observations and leverage the spatial correlation features from neighboring sensors to predict missing values. Indeed, it is crucial to emphasize that sensor \#2, despite lacking observed values for two hours and thus not having access to its temporal correlation information, can still rely on observed data from spatial neighboring sensors to estimate the missing values. This highlights the capability of our FastSTI to achieve relatively satisfactory imputation results.

\section{Conclusion and Future Work} \label{sec:6}
In this paper, we investigate the challenges of spatiotemporal traffic data imputation in real-life scenarios, proposing our FastSTI model. FastSTI utilizes a higher-order pseudo-numerical methodology for a conditional diffusion model to enhance imputation accuracy. Furthermore, we demonstrate the effectiveness of incorporating conditional information through GCN variants (Diff-GCN) to serve as feature prior knowledge, capturing the spatiotemporal correlations. To accelerate the imputation process and fit real-life applications, we introduce the utilization of a variance schedule to reduce the sampling iterations of the diffusion model. The effectiveness of the proposed FastSTI in addressing real-world data missing scenarios is substantiated by experiments, showcasing its accelerated imputation time and competitive performance in imputation tasks.

In FastSTI, our primary focus is on a single feature attribute: traffic speed or flow. Traffic conditions are also influenced by various other factors, such as weather, Points of Interest (POI), and unexpected events. More attributes can be considered in the process of missing data estimation. Exploring this idea further could be a notable topic for future researchers.

\section*{Acknowledgments}
This work was supported in part by China Scholarship Council under Grant 202206290090, in part by the University of Padova (UniPD) Budget Integrato per la Ricerca dei Dipartimenti (BIRD)-2021 Project, and in part by the Progetti di Ricerca di Interesse Nazionale (PRIN)-17 PREVUE Project from Italian Ministry for Universities and Research (MUR) under Grant CUP E94I19000650001. 

\bibliographystyle{IEEEtran}
\bibliography{refs}

\vfill

\end{document}